\documentclass{article}

\usepackage{microtype}
\usepackage{graphicx}
\usepackage{subcaption}
\usepackage{booktabs}

\usepackage[preprint,nohyperref]{icml2026}

\usepackage{amsmath}
\usepackage{amssymb}
\usepackage{mathtools}
\usepackage{amsthm}

\usepackage{url}
\usepackage[most]{tcolorbox}
\usepackage{tikz}
\usepackage{color, colortbl}
\usepackage[utf8]{inputenc} % allow utf-8 input
\usepackage[T1]{fontenc}    % use 8-bit T1 fonts
\usepackage{amsfonts}       % blackboard math symbols
\usepackage{nicefrac}       % compact symbols for 1/2, etc.
\usepackage{xcolor}         % colors
\usepackage{siunitx}
\tcbset{
  promptbox/.style={
    breakable,
    enhanced,
    colback=gray!3,
    colframe=gray!55,
    coltitle=black,
    fonttitle=\bfseries,
    title={#1},
    boxrule=0.6pt,
    arc=2mm,
    left=1.2ex, right=1.2ex, top=1ex, bottom=1ex,
    before skip=6pt, after skip=8pt
  }
}

\usepackage[capitalize,noabbrev]{cleveref}

\theoremstyle{plain}

\theoremstyle{definition}

\theoremstyle{remark}

\usepackage[textsize=tiny]{todonotes}

\icmltitlerunning{Towards Understanding Multimodal Fine-Tuning: Spatial Features}

\begin{document}

\twocolumn[
  \icmltitle{Towards Understanding Multimodal Fine-Tuning: A Case Study into Spatial Features}

  % Authors (ICML format) — same content as your ICLR author block
  \begin{icmlauthorlist}
    \icmlauthor{Lachin Naghashyar}{oxford}
    \icmlauthor{Hunar Batra}{oxford}
    \icmlauthor{Ashkan Khakzar}{oxford}
    \icmlauthor{Philip Torr}{oxford}
    \icmlauthor{Ronald Clark}{oxford}
    \icmlauthor{Christian Schroeder de Witt}{oxford}
    \icmlauthor{Constantin Venhoff}{oxford}
  \end{icmlauthorlist}

  \icmlaffiliation{oxford}{University of Oxford, Oxford, United Kingdom}

  \icmlcorrespondingauthor{Lachin Naghashyar}{lachin.naghashyar@cs.ox.ac.uk}

  \icmlkeywords{Machine Learning, ICML}

  \vskip 0.3in
]

\printAffiliationsAndNotice{}  % no special notice (required even if empty)

%%%%%%%%%%%%%%%%%%%%%%%%%%%%%%%%%%%%%%%%%%%%%%%%%%%%%%%%%%%%%%%%%%%%%%%%%%%%%%%%%%
\begin{abstract}
Contemporary Vision–Language Models (VLMs) achieve strong performance on a wide range of tasks by pairing a vision encoder with a pre-trained language model, fine-tuned for visual–text inputs. Yet despite these gains, it remains unclear how language backbone representations adapt during multimodal training and when vision-specific capabilities emerge. In this work, we present the first mechanistic analysis of VLMs adaptation process. Using stage-wise model diffing, a technique that isolates representational changes introduced during multimodal fine-tuning, we reveal how a language model learns to "see". We first identify vision-preferring features that emerge or reorient during fine-tuning. We then show that a selective subset of these features reliably encodes spatial relations, revealed through controlled shifts to spatial prompts. Finally, we trace the causal activation of these features to a small group of attention heads. Our findings show that stage-wise model diffing reveals when and where spatially-grounded multimodal features arise. It also provides a clearer view of modality fusion by showing how visual grounding reshapes features that were previously text-only. This methodology enhances the interpretability of multimodal training and provides a foundation for understanding and refining how pretrained language models acquire vision-grounded capabilities.

% feedback: Do we mention how? if not, better to remove and be to the point -> agreed

% This methodology enhances the interpretability of multimodal training and provides a foundation for refining training regimes as well as auditing and steering models in safety-critical or domain-specific settings.

\end{abstract}
%%%%%%%%%%%%%%%%%%%%%%%%%%%%%%%%%%%%%%%%%%%%%%%%%%%%%%%%%%%%%%%%%%%%%%%%%%%%%%%%%

%%%%%%%%%%%%%%%%%%%%%%%%%%%%%%%%%%%%%%%%%%%%%%%%%%%%%%%%%%%%%%%%%%%%%%%%%%%%%%%%%%%
\section{Introduction}
% Figure: Overall figure. 1) Depict SAE fine tuning. 2) Depict spatial feature search. 3) Depict spatial attention heads

Large vision–language models (VLMs) have achieved strong performance on multimodal tasks, including visual question answering (VQA), image captioning, object detection, and visual grounding \citep{li2024llava, mistral2024pixtral}. These gains are typically realized by fine-tuning pretrained language models to process visual inputs through projected token sequences, allowing for seamless fusion of image and text representations \citep{xu2024llava, zhang2024improve, dong2024insight, Dong2024ProgressiveMR}. Yet we lack a mechanistic account of how language representations adapt during multimodal training and when vision-specific capabilities emerge \citep{khayatan2025analyzing, venhoff2025visual, stan2024lvlm}.

In this work, we introduce a method for analyzing multimodal adaptation in VLMs through stage-wise model diffing \citep{Bricken2024StageWiseModelDiffing}. This mechanistic interpretability technique isolates representational changes introduced during fine-tuning by comparing sparse autoencoder (SAE) dictionaries across training stages, models, or datasets. By tracking how features rotate, emerge, or are repurposed, it has been shown to uncover subtle shifts such as sleeper-agent features \citep{hubinger2024sleeper}. We extend this approach to the multimodal setting, presenting the first application of stage-wise model diffing to study how pretrained language features evolve under visual grounding.

Concretely, we fine-tune LLaMA-Scope SAEs on activations extracted from the LLaVA-More model \citep{he2024llama} on 50k VQAv2 dataset samples \citep{goyal2017making}. This warm-start preserves the original feature basis while adapting to multimodal activations. We isolate features that gain visual preference and undergo strong geometric rotation, serving as anchors for studying spatial representations in the backbone. To identify which adapted features encode spatial reasoning, we apply a controlled dataset shift from general VQA to spatial queries. Features that are preferentially recruited under spatial prompts form a selective subset, which we validate through automatic and manual interpretation. These features consistently activate on questions about object placement, relative position, and orientation. Figure~\ref{fig:global-scatter} highlights the filtered spatial features.

Finally, we use attribution patching to trace the causal pathways by which these spatial features are activated. Our results reveal a sparse set of mid-layer heads that consistently drive spatial representations, often localizing to semantically meaningful regions and reappearing across related prompts. These findings support the hypothesis that a small number of specialized attention heads coordinate visual grounding within the model. Our contributions are as follows:
\begin{itemize}
 \item We extend stage-wise model diffing to the multimodal setting, providing the first feature-level account of how pretrained language backbones adapt under visual grounding.
\item We introduce a systematic pipeline to isolate adapted features, identify those selectively recruited by spatial queries, and filter out lexical artifacts.
\item We show that these spatially selective SAE features are functionally involved in reasoning, through empirical evidence and ablation studies, supported by interpretive checks.
\item We causally attribute the emergence of spatial features to a small subset of attention heads using scalable attribution patching, highlighting structured pathways for visual grounding.
\end{itemize}

By focusing on feature-level change, our approach complements high-level alignment analyses and probing-based methods, providing a deeper mechanistic view of how models “learn to see”. More broadly, this work offers a framework for auditing and refining multimodal training regimes, with implications for safety-critical domains and targeted fine-tuning in specialized applications.

%%%%%%%%%%%%%%%%%%%%%%%%%%%%%%%%%%%%%%%%%%%%%%%%%%%%%%%%%%%%%%%%%%%%%%%%%%%%%%%%%%%%%

%%%%%%%%%%%%%%%%%%%%%%%%%%%%%%%%%%%%%%%%%%%%%%%%%%%%%%%%%%%%%%%%%%%%%%%%%%%%%%%%%%%%%
\section{Related Work}

\paragraph{Model Diffing and Representation Dynamics} Model diffing techniques aim to isolate how internal representations change across models or training stages. 
Early work focused on coarse similarity measures, such as visualizing function-space geometry \citep{olah2015visualizing, erhan2010}, stitching intermediate layers across models \citep{lenc2015understanding, bansal2021revisiting}, or defining new similarity metrics \citep{kornblith2019similarity, barannikov2021representation}. 
Later studies examined alignment at the level of individual neurons, showing convergent units across independently trained networks \citep{li2015convergent, olah2020zoom}. 

Sparse autoencoders (SAEs) offered a feature-level lens, and prior work \citep{kissane2024saes} showed that SAEs largely transfer between base and fine-tuned models, implying most features are preserved and only a minority are altered. This motivates methods that can isolate and precisely interpret those changes. Stage-wise model diffing \citep{Bricken2024StageWiseModelDiffing} offers such fine-grained resolution, revealing sleeper-agent features and distinguishing between base and chat-tuned models \citep{minder2025robustly}. Extensions to multimodal models highlight similar representational shifts, with concept-shift vectors proposed for steering \citep{khayatan2025analyzing} and evidence that alignment converges in middle-to-late layers \citep{venhoff2025visual}. These remain semantic-level analyses, whereas our work applies stage-wise diffing with SAEs to the backbone, giving the first mechanistic account of multimodal fine-tuning, showing how it rotates features and induces spatial grounding in pretrained language models.

\paragraph{Multimodal Mechanistic Interpretability.}
Compared to the rapidly growing literature on mechanistic interpretability of textual LLMs, relatively few studies have examined the internal mechanisms of multimodal large language models (MLLMs). Existing work falls into two main categories. 

First, tool-based and causal analyses aim to explain model behavior at a high level. Approaches include interpretability toolkits based on attention patterns, relevancy maps, and causal interventions \citep{stan2024lvlm}. Other work uses interventions to trace how information is stored and transferred \citep{basu2024understanding}, or applies causal mediation to study how BLIP integrates visual evidence \citep{palit2023towards}. Second, probing-based studies focus on the representations themselves. Several works analyzed CLIP, identifying both strengths and limitations \citep{tong2024eyes, gandelsman2023interpreting, chen2023interpreting}. Others reported multimodal neurons responsive to joint visual–textual concepts \citep{schwettmann2023multimodal} and examined how VLMs differentiate hallucinated from real objects \citep{jiang2024interpreting}. More recent methods map visual embeddings into linguistic space, projecting features onto language vocabularies \citep{neo2024towards} or showing the late emergence of visual signals in LLM backbones \citep{venhoff2025toolate}. 

In contrast, these studies primarily analyze patterns, interventions, or probing correlations, but do not directly track how multimodal fine-tuning restructures the backbone’s internal features. Our work addresses this gap by providing a mechanistic perspective.

%%%%%%%%%%%%%%%%%%%%%%%%%%%%%%%%%%%%%%%%%%%%%%%%%%%%%%%%%%%%%%%%%%%%%%%%%%%%%%%%%%%%%
\section{Preliminaries}
\subsection{Vision--Language Models}

A vision--language model (VLM) consists of a visual encoder $f_V$, a pretrained language model $f_{\mathrm{LM}}$, and a trainable projector $P$. The visual encoder (e.g., a ViT \citep{radford2021learning}) extracts image patch embeddings $V = f_V(x) = [v_1, \ldots, v_{N_V}]$, which the projector maps into token space $\tilde{V} = P(V)$. These projected image tokens are concatenated with tokenized text embeddings $T = [t_1, \ldots, t_{N_T}]$ to form the multimodal sequence $X = [\tilde{v}_1, \ldots, \tilde{v}_{N_V}, t_1, \ldots, t_{N_T}]$. Alignment between modalities is achieved through \emph{visual instruction tuning}, where image–text pairs fine-tune the backbone to follow multimodal instructions. The language model processes $X$ through transformer layers of multi-head self-attention and feed-forward networks. For each head $h$, attention is computed as
\begin{equation}
    \text{Attn}(Q,K,V) = \text{Softmax}\!\Big(\tfrac{QK^\top}{\sqrt{d_h}} + M\Big)V,
\end{equation}
where $M$ is the causal mask preventing attention to future tokens. The outputs of all heads are concatenated and projected into the hidden dimension, and mapped through the unembedding matrix to predict next tokens.  For our experiments, we adopt LLaVA-More \citep{cocchi2025llava}, which extends LLaVA framework \citep{liu2023visual, liu2024improved} by integrating recent language models and diverse visual backbones; specifically, we use the variant combining the CLIP ViT-Large-Patch14--336 encoder with a LLaMA-3.1-8B language model backbone \citep{grattafiori2024llama}. 

\subsection{Sparse Autoencoders (SAEs)}

Sparse Autoencoders (SAEs) learn a dictionary of features that approximate hidden states as sparse linear combinations of interpretable directions. mitigating superposition where many features overlap in the same dimensions \citep{bricken2023monosemanticity, cunningham2023sparse}. Formally, a vanilla SAE encodes $x \in \mathbb{R}^D$ into
\[
f(x) = \text{ReLU}(W_{\text{enc}}x + b_{\text{enc}}), \quad 
\hat{x} = W_{\text{dec}} f(x) + b_{\text{dec}},
\]
with $W_{\text{enc}} \in \mathbb{R}^{F \times D}$, $b_{\text{enc}} \in \mathbb{R}^F$, 
$W_{\text{dec}} \in \mathbb{R}^{D \times F}$, and $b_{\text{dec}} \in \mathbb{R}^D$. 
Training minimizes
\[
\mathcal{L} = \|x - \hat{x}\|_2^2 + \lambda \sum_{i=1}^F | f_i(x) |,
\]

combining reconstruction with an $L_1$ sparsity penalty. Here, decoder columns $(W_{\text{dec}})_{:,i}$ define the direction of each feature in input space, while encoder rows $(W_{\text{enc}})_{i,:}$ act as detectors that determine when a feature is present. Variants such as Top-$K$ SAEs \citep{gao2024scaling} further sharpen this tradeoff by enforcing hard sparsity, improving interpretability and reducing feature co-adaptation.

SAEs have been widely applied to uncover monosemantic features and offer a practical lens on model internals, enabling analyses that range from probing knowledge to tracing safety-relevant behaviors \citep{bricken2023monosemanticity, cunningham2023sparse}. They are not, however, a complete decomposition: interpretability can vary across runs and training setups, and recent work suggests their practical utility may be more limited in some settings \citep{templeton2024scaling, kantamneni2025saeprobes}. Even so, SAEs have proven particularly effective for \emph{model diffing}, where they make it possible to track how features shift across training stages and to surface subtle but behaviorally important dynamics—a direction we expand on in the next subsection \citep{Bricken2024StageWiseModelDiffing, minder2025diff}.

\subsection{Stage-Wise Model Diffing}
\label{sec:diffing_background}

A recent line of work in model diffing has introduced \emph{stage-wise model diffing}~\citep{Bricken2024StageWiseModelDiffing}, which extends SAE analysis across training stages by re-training dictionaries on activations from successive checkpoints while keeping feature indices aligned. This makes it possible to compare whether units are preserved, rotated, or repurposed during adaptation. Applied to controlled fine-tuning trajectories, it disentangles changes due to model updates from dataset shifts and highlights features that drive adaptation. Prior work has shown that stage-wise diffing uncovers fine-grained dynamics, including sleeper-agent features that remain dormant in pretraining but activate once safety constraints are lifted~\citep{hubinger2024sleeper}. Compared to crosscoder-based methods \citep{lindsey2024crosscoders}, it provides finer resolution at the feature level, though it remains limited to aligned checkpoints of the same architecture.

%≈

%%%%%%%%%%%%%%%%%%%%%%%%%%%%%%%%%%%%%%%%%%%%%%%%%%%%%%%%%%%%%%%%%%%%%%%%%%%%%%%%%%%%%%%

\section{Stage-wise Model Diffing for Multimodal Adaptation}
\label{sec:method-overview}

\paragraph{Overview.}

We aim to understand how multimodal fine-tuning reshapes model representations, using spatial
reasoning as a case study of a distinctly multimodal task that integrates both visual and linguistic
cues. To this end, we take inspiration from stage-wise diffing~\ref{sec:diffing_background},
employing sparse autoencoders (SAEs) as a feature-level lens to track how internal directions shift
when a pretrained language backbone is exposed to visual inputs. Our pipeline has three stages.
First, we fine-tune SAEs on multimodal activations to obtain a feature dictionary aligned with the
vision–language space. Second, we isolate features that prefer visual tokens and undergo substantial
geometric rotation, indicating that they have been repurposed by multimodal training. Third, we probe
for spatial reasoning by contrasting generic VQA with spatial queries and keeping only features that
increase under the shift while remaining active under neutral instructions, ensuring they are not driven
by lexical artifacts. In this way, we reduce the original pool of over one million features to a compact
set of candidates plausibly recruited for spatial reasoning tasks.
\subsection{Adapting Language Dictionaries to Vision-Language Space}
\label{sec:sae-finetune}

We start by adapting sparse autoencoders (SAEs) trained on the Llama 3.1 8B backbone to the hidden states of \textsc{LLaVA-MORE} (Llama 3.1 8B backbone) \citep{cocchi2025llava}. We use 50k image–question pairs from the VQAv2 dataset \citep{goyal2017making}, a widely used VQA benchmark of images and open-ended questions. Each SAE is attached to the output of a transformer block and trained on cached activations from these samples. Images are represented by 575 consecutive visual tokens, and questions by variable-length text sequences; this separation allows token-type–specific masking.

We initialize SAEs from the pretrained \textsc{LLaMA-Scope} release \citep{he2024llama}, re-instantiated as a Top-$K$ model ($k{=}50$), preserving a meaningful, interpretable basis. Since our VLM shares the same backbone, this warm-start ensures continuity with the pretrained language feature space and avoids retraining from scratch, allowing us to directly leverage millions of monosemantic features across layers. As a control, we also train SAEs from random initialization under identical conditions. Training uses Adam with a layer-scaled learning rate, and cached activations are processed in padded mini-batches. To disentangle modality-specific contributions, we consider four regimes: (i) full sequence, (ii) image-only, using only the visual-token span, (iii) text-only, using only the non-visual span, and (iv) random initialization. In all cases, the SAE receives the full hidden state sequence, but masking controls which token spans contribute to the training signal.

\begin{figure}[h]
  \centering

  % --- Top: plot ---
  \includegraphics[width=0.9\linewidth,clip,trim=0 0 0 8]{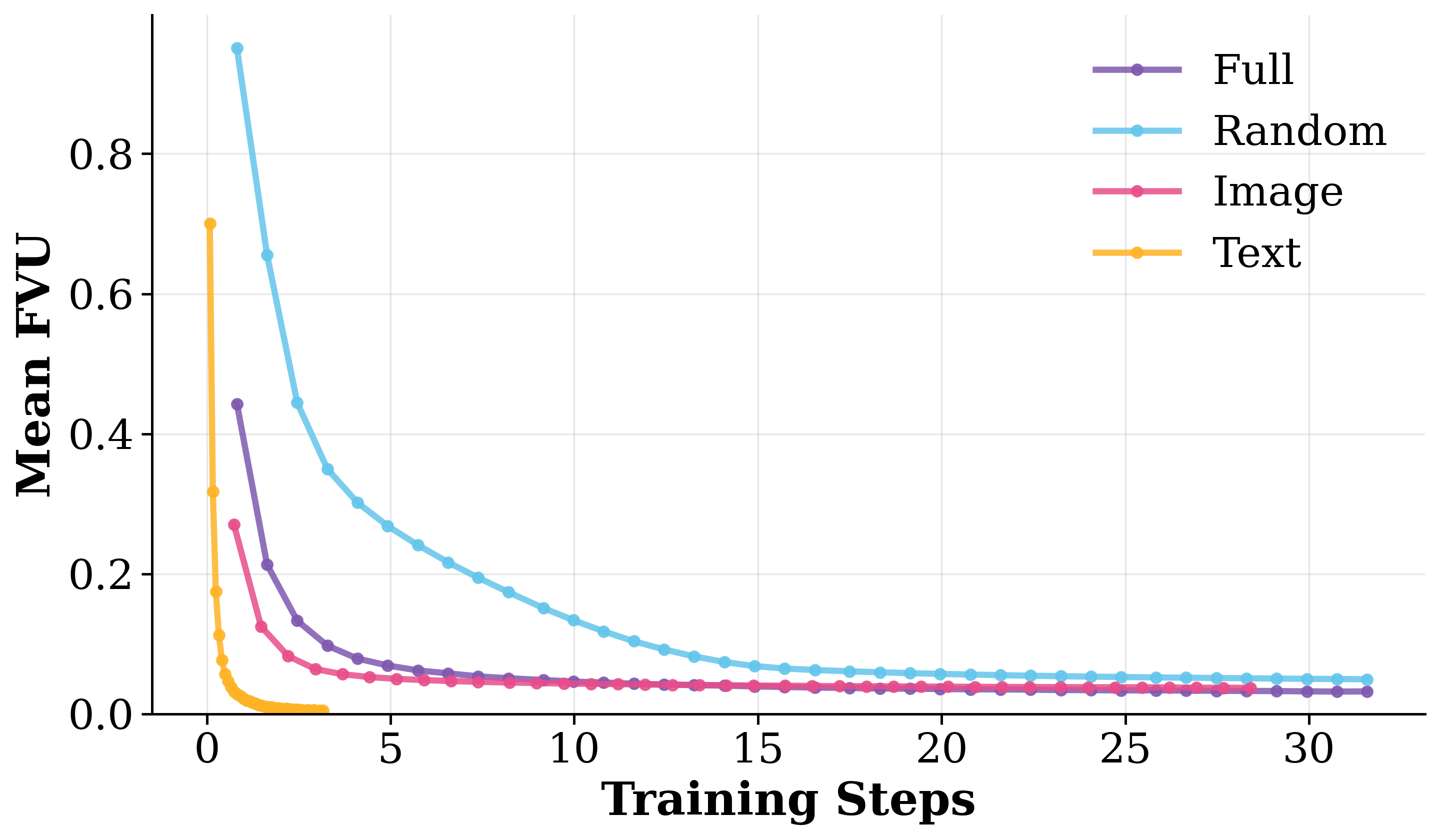}

  \vspace{0.9em} % space between plot and table

  % --- Bottom: table ---
  \scriptsize
  \setlength{\tabcolsep}{4pt}
  \renewcommand{\arraystretch}{1.05}
  \resizebox{\linewidth}{!}{%
    \begin{tabular}{
      l
      S[table-format=2.3]
      S[table-format=2.3]
      S[table-format=2.3]
      S[table-format=2.3]
    }
      \toprule
      \textbf{Metric} & {Full} & {Random} & {Image} & {Text} \\
      \midrule
      Mean        & 0.032 & 0.050 & 0.037 & \bfseries 0.005 \\
      Std         & 0.028 & 0.041 & 0.027 & \bfseries 0.009 \\
      Min         & 0.013 & 0.020 & 0.017 & \bfseries 0.000 \\
      Max         & 0.123 & 0.198 & 0.123 & \bfseries 0.037 \\
      {\tiny Tokens (M)} & {\tiny 31.6} & {\tiny 31.6} & {\tiny 28.4} & {\tiny 3.2} \\
      \bottomrule
    \end{tabular}
  }

  \caption{\textbf{SAE adaptation on \textsc{LLaVA-MORE}.}
  Top: Mean fraction of variance unexplained (FVU) across layers on the validation set.
  Bottom: Summary statistics of FVU values on the validation set, with decimal alignment; the lowest mean is highlighted in \textbf{bold}.}

  \label{fig:fvu-curves}
\end{figure}

We evaluate reconstruction quality using the fraction of variance unexplained (FVU) and report sparsity to verify code selectivity. Evaluation is performed on a held-out split. Figure \ref{fig:fvu-curves} shows FVU as a function of tokens seen across layers and masking regimes. 
Text-only SAEs converge rapidly, while image-only and full-token regimes converge more slowly to higher error, reflecting the mismatch between projector embeddings and the LLM basis. 
Random initialization performs worst, underscoring the importance of starting from a pretrained language dictionary. 
These findings establish text-only SAEs as a reliable reconstruction baseline, which we later use for model diffing.

\paragraph{Implications for stage-wise model diffing.}
Stage-wise diffing assumes that fine-tuning induces \emph{localized} (feature-level) changes rather than wholesale rotations.
Prior work reports that image-token representations in early layers exhibit higher reconstruction error than text tokens, indicating a distributional gap between projector outputs and the LLM basis \citep{venhoff2025twohop}.
Consistent with this, our decoder–cosine analysis (Appx.Fig.\ref{fig:decoder-cos-trends}) shows that \emph{text-only} SAEs remain highly aligned to the base LLM dictionary across layers, whereas \emph{image-only} and \emph{full sequence} SAEs undergo large rotations in shallow layers and only align in later layers.
We also note that text-only SAEs begin with slightly higher error in the very first layers but adapt extremely quickly, converging to near-zero reconstruction. In contrast, image and full-sequence SAEs plateau at higher error, highlighting the instability of projector-driven spans (see Appx.Fig.\ref{fig:fvu-layerwise}).
We therefore focus stage-wise diffing on text-only SAEs, where alignment is stable and feature-level identifiability is more plausible.

%%%%%%%%%%%%%%%%%%%%%%%%%%%%%%%%%%%%%%%%%%%%%%%%%%%%%%%%%%%%%%%%%%%%%%%%%%%%%
% --- Section 4: Identifying Candidate Features (matched to Section 3 style) ---
\subsection{Identifying Adapted Features}
\label{sec:adapted}

We aim to isolate SAE features that (i) undergo geometric reorientation after multimodal adaptation and (ii) show a clear \emph{modality preference} for vision input. Such features are the most informative for model diffing and subsequent causal analysis. To identify them, we rely on two signals:

\textbf{1. Geometric reorientation (decoder cosine).}
To test if $f$ has been \emph{repurposed} by multimodal fine-tuning, we compare its decoder direction before and after adaptation. Let $W^{\text{LLM}}_{\text{dec},f}$ be the base SAE decoder vector and $W^{\text{VLM}}_{\text{dec},f}$ the corresponding vector in the VLM-adapted SAE. We compute
\[
c_f \;=\; \cos\!\bigl(W^{\text{LLM}}_{\text{dec},f},\, W^{\text{VLM}}_{\text{dec},f}\bigr).
\]
High $c_f$ means the semantic direction of $f$ stayed aligned with the original language dictionary; low $c_f$ indicates a substantial rotation, consistent with a reallocation of $f$ to encode new multimodal structure.
We use decoder vectors rather than encoder parameters because decoder directions more directly index the feature’s semantics.

\textbf{2. Modality preference (visual energy).}
Given the sparsity of SAE activations, we score each feature $f$ by its mean squared activation under vision inputs,
\[
E_v(f) \;=\; \mathbb{E}_{\text{vision}}\!\big[h_f^2\big],
\]
measured on VQA runs of the VLM. Since nearly half of features have $E_v=0$, a simple cutoff $E_v>\epsilon$ suffices to discard inactive directions and retain those that carry visual signal.

\paragraph{Selection Procedure}

We define adapted features as those that meet both criteria: $E_v>\epsilon$, ensuring reliable visual responsiveness, and a cosine similarity $c_f$ in the bottom $p_{cos}=25\%$, indicating strong decoder rotation. Applying these filters jointly yields a globally defined set comprising about $5\%$ of all features. The joint distribution of $E_v$ and $c_f$ is shown in Fig.~\ref{fig:global-scatter}, with the selected subset highlighted in pink. Details on threshold choices, together with per-layer counts and mean cosine similarities, are provided in Appx.~Fig.~\ref{fig:suspects-per-layer} and Appx.~\ref{fig:cosine-overall-vs-suspect}.

% Figure: global joint distribution with highlights

%%%%%%%%%%%%%%%%%%%%%%%%%%%%%%%%%%%%%%%%%%%%%%%%%%%%%%%%%%%%%%%%%%%%%%%%%%%%%%%%

\subsection{Case Study: Identifying Spatial Reasoning Features}
\label{sec:spatial}

We identify spatial features using two signals: (i) recruitment under a shift to spatial queries, and (ii) persistence under neutral prompts that rule out lexical artifacts.

\paragraph{Datasets.}
Our analysis uses two evaluation sets from VQAv2. The baseline is the full validation split, denoted $\mathcal{D}_{\text{base}}$. To induce a targeted shift, we construct a spatial subset $\mathcal{D}_{\text{sp}}$ by filtering questions that contain spatial cues (e.g., \emph{left/right/above/behind}). This contrast tests whether some SAE features are selectively recruited under spatial reasoning.

\begin{figure*}[t]
  \centering
  \includegraphics[width=0.95\textwidth]{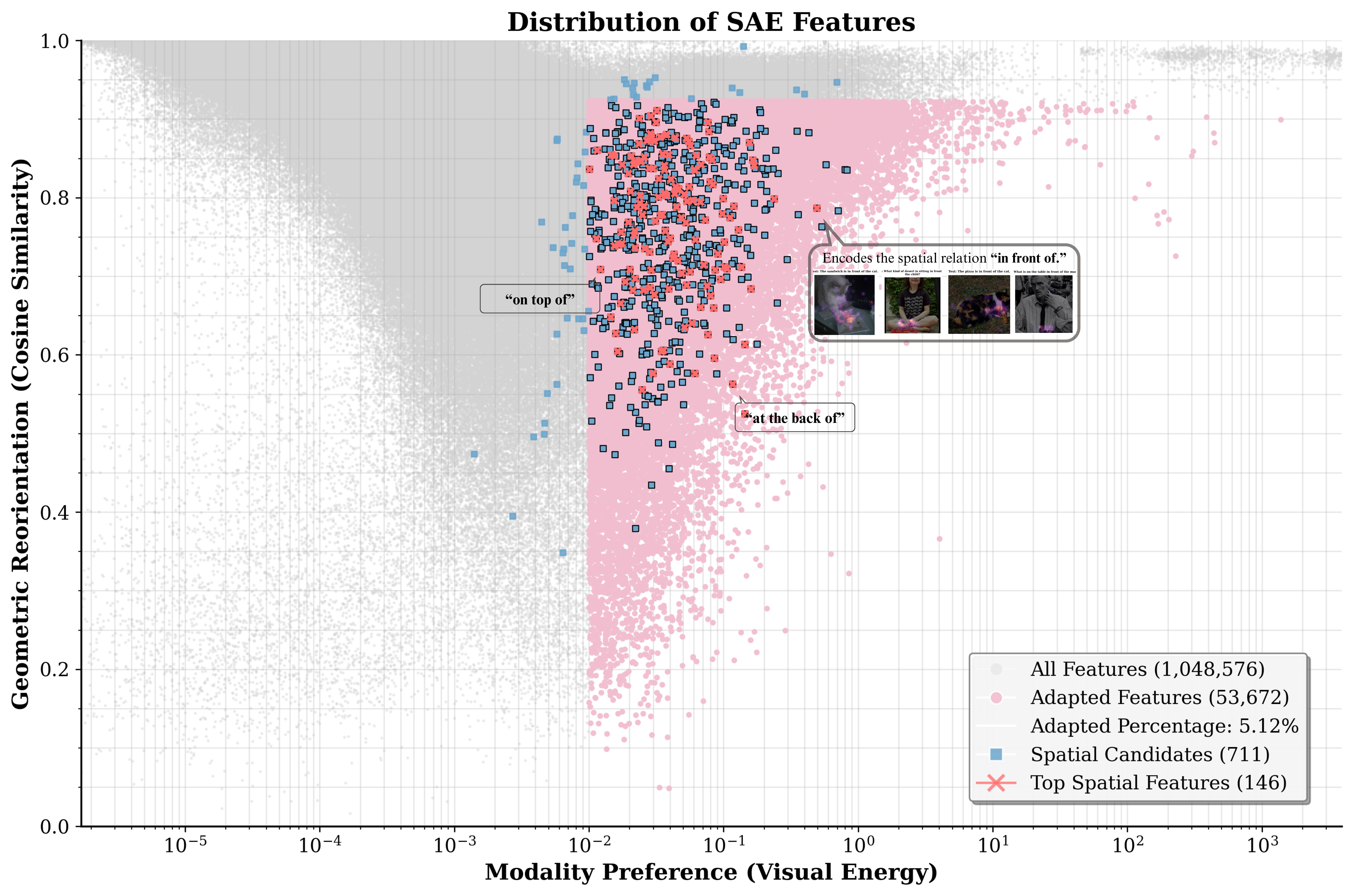}

  \caption{\textbf{Distribution of SAE features by visual energy and cosine similarity.} 
  All features are shown in gray; adapted features are highlighted in pink. 
  Spatial candidates are marked with blue squares, and the subset used for downstream analysis is shown as red crosses.}

  \label{fig:global-scatter}
\end{figure*}

\paragraph{1. Distribution shift}
Let $h_f(x_t) \ge 0$ denote the activation of feature $f$ on token $t$ of input $x$. For a dataset $\mathcal{D}$, the firing frequency of $f$ is
\[
p_f(\mathcal{D}) \;=\; \frac{1}{n(\mathcal{D})}\sum_{x\in\mathcal{D}}\sum_t \mathbf{1}\{h_f(x_t)>0\},
\]
where $n(\mathcal{D})$ is the total number of tokens.
We compute this measure for the base split $\mathcal{D}_{\text{base}}$ and a spatial split
$\mathcal{D}_{\text{sp}}$, and evaluate each feature using the frequency gap
\(
\Delta p_f = p_f(\mathcal{D}_{\text{sp}}) - p_f(\mathcal{D}_{\text{base}})
\)
alongside its odds ratio $\mathrm{OR}_f$. Features with meaningful $\Delta p_f$ and $\mathrm{OR}_f$ are flagged as spatial \emph{candidates} in Fig.\ref{fig:global-scatter}. Further details, including firing-frequency and scatter-plot visualizations for both splits, are provided in Appx.\,\ref{app:dist-shift}.

\paragraph{2. Filtering lexical artifacts.}
To rule out prompt-lexical effects, we replace the original questions in each top-activating sample with neutral spatial prompts such as \emph{``Describe the positions of objects in the image.''}. Features that continue firing under these generic instructions are preserved as genuinely image-grounded, while those that fail to activate are discarded. This ensures that the surviving units reflect spatial reasoning rather than memorized lexical cues.

From these filtered candidates, we retain only those also in the adapted set $\mathcal{A}$ (Sec.~\ref{sec:adapted}), ensuring they reorient under multimodal fine-tuning and respond to spatial shifts. The surviving features are shown in Fig.~\ref{fig:global-scatter} (blue). A subset, marked with red crosses, is further analyzed via automated interpretation, attribution patching, and ablations (Sec.~\ref{sec:auto-interp}, \ref{sec:attr-patch}).

\paragraph{Extension to OCR-style prompts.}
While our primary case study focuses on spatial reasoning, the same feature-selection procedure
can be applied to other visually grounded skills. As a second case study, we analyze features
associated with visual text recognition by contrasting OCR-style prompts (e.g., ``What does the
sign say?'') with generic VQA questions. We construct an OCR-focused split by filtering VQAv2
images that contain legible embedded text and computing feature firing frequencies under the same
distribution-shift statistics used for spatial queries. This reveals a compact subset of adapted units
whose activations increase on OCR prompts and remain non-zero under neutral image descriptions,
indicating that they are tied to image-grounded text rather than specific lexical patterns. Additional
qualitative examples and follow-up analyses are provided in Appx.~\ref{app:ocr}, where these
OCR-selective features are shown to align with regions containing characters and words and to be
supported by a small number of recurring mid-layer heads.

\section{Experiments}
\subsection{Auto-Interp and Preliminary Inspection}
\label{sec:auto-interp}

\begin{figure*}[t]
  \centering
  \includegraphics[width=\linewidth]{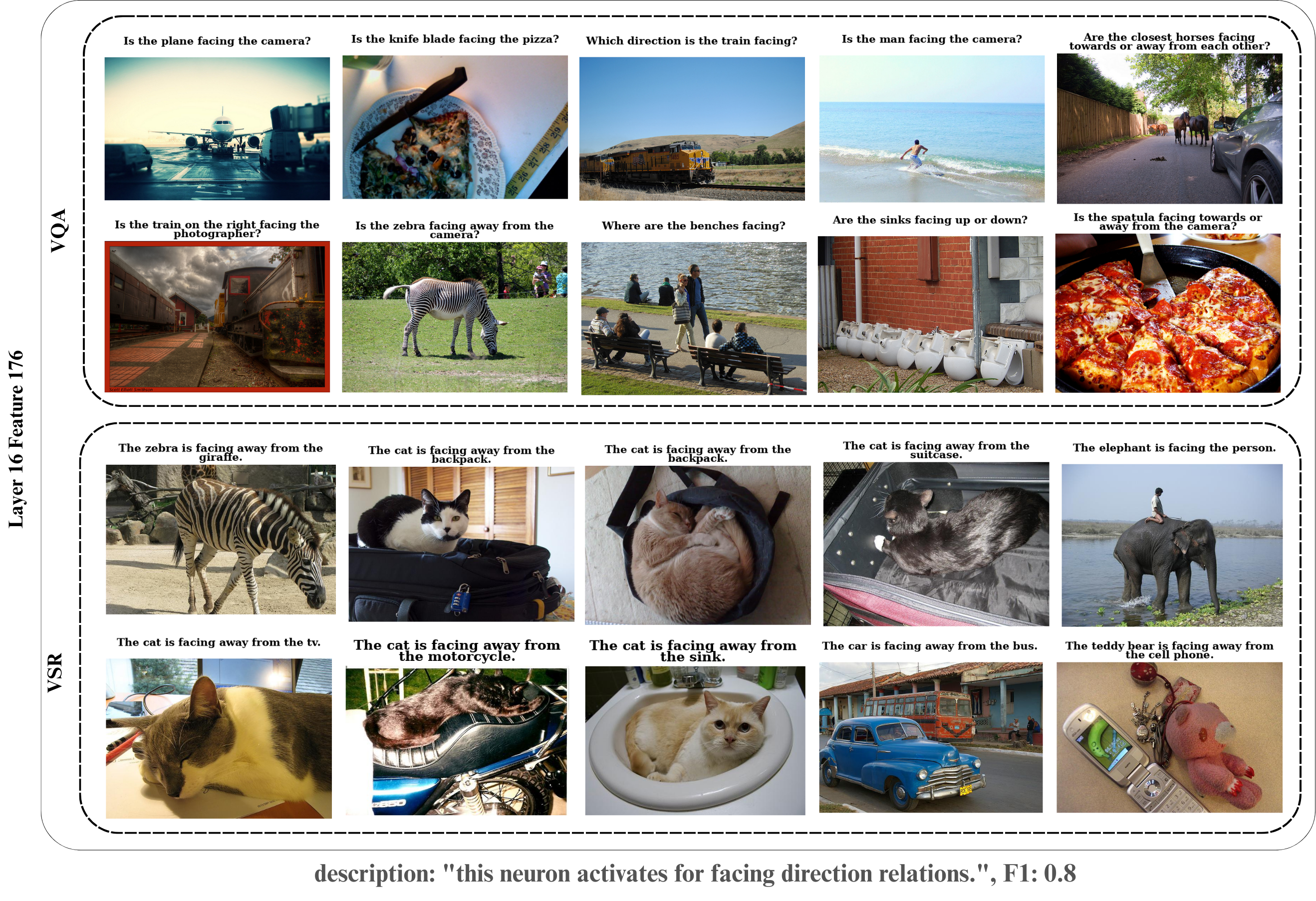}
  \vspace{-1em}
  \caption{\textbf{Auto-Interp example (Layer 16, Feature 176).} 
Top VQA and VSR samples both highlight \emph{facing direction}, with activation on objects described as facing toward, away, or relative to another.}
  \label{fig:auto-interp-example}
\end{figure*}

As an initial step toward understanding the selected features, we carried out a preliminary inspection using an automated interpretation pipeline. For each feature, we collect its top-activating samples from two sources: general VQA questions from VQAv2 (not restricted to spatial reasoning) and the Visual Spatial Reasoning (VSR) dataset~\citep{liu2023vsr}, which is inherently spatial. This pairing allows us to check whether the same underlying meaning
emerges consistently across both settings (Fig.~\ref{fig:auto-interp-example}). A subset of the combined samples are then passed to the \texttt{gpt-4o-mini}~\citep{gpt-4o-mini} API, which proposes a concise one-sentence description for each feature and assigns an interpretability confidence score based on F1 from a validation classification task. The resulting outputs are stored together with the selection metrics from Sec.~\ref{sec:adapted}, and are lightly reviewed by hand, so that the retained set reflects both
automatic labeling and human verification (see App.~\ref{app:auto-interp} for additional examples and scoring details)..

%%%%%%%%%%%%%%%%%%%%%%%%%%%%%%%%%%%%%%%%%%%%%%%%%%%%%%%%%%%%%%%%%%%%%%%%%%%%%%%%%%%%%%%%%%%

\vspace{-1em}

\subsection{Attribution Patching to Identify Spatial Heads}
\label{sec:attr-patch}

\paragraph{Method.}
Attribution patching \citep{nanda2023attribution} is a scalable alternative to activation patching \citep{zhang2024activationpatching}, which measures causal effects by replacing activations with counterfactuals. While activation patching requires a separate forward pass per intervention, attribution patching uses a gradient-based linear approximation to estimate interventions with two forward and one backward pass. This makes it practical to probe attribution scores across layers and heads in MLLMs.

We adapt attribution patching to identify which attention heads drive spatially selective SAE features. 
For a target feature $f$ at layer $L$, we define a scalar objective by projecting the layer-$L$ activations onto the SAE decoder vector.
Gradients of this objective w.r.t. upstream query/key activations indicate how strongly each attention head contributes to $f$. We compare two runs:
\begin{itemize}
    \item \textbf{Clean run:} the original image--text input. 
    \item \textbf{Corrupt run:} the same input, but with layer-0 visual token embeddings replaced by a \emph{mean embedding} computed over many VQA samples. This corruption preserves plausible distributional statistics while deliberately suppressing spatial information.
\end{itemize}

We then compute two attribution variants, differing in whether the perturbation direction is taken from the corrupted or the clean representation:
\begin{align*}
\text{Method A:} \quad & (\text{corr} - \text{clean}) \cdot \nabla_{\text{clean}}, \\
\text{Method B:} \quad & (\text{clean} - \text{corr}) \cdot \nabla_{\text{corr}} .
\end{align*}
Method~A measures how strongly the clean gradients indicate that ablating spatial detail affects the feature, whereas Method~B measures how strongly the corrupted gradients indicate that retaining spatial detail matters. 
In both cases, we obtain per-layer and per-head attribution scores, averaged over the top-$k$ VQA samples that most strongly activate $f$.

\paragraph{Results.}
Across the spatially selective features we examined, attribution patching with both methods reveals consistent trends. Layer-wise attribution curves typically peak in middle layers, consistent with the emergence of spatial features in Sec.~\ref{sec:spatial} (Appx.~Fig.~\ref{fig:layer-agg}). At the head level, both methods generally highlight a small subset of heads with notably high scores, and the top heads identified are often consistent across the two attribution methods (Appx.~Fig.~\ref{fig:head-agg}). This suggests that spatial information is mediated by a specialized group of heads rather than being spread uniformly across the model.

\begin{figure*}[t]
  \centering
  \includegraphics[width=\linewidth]{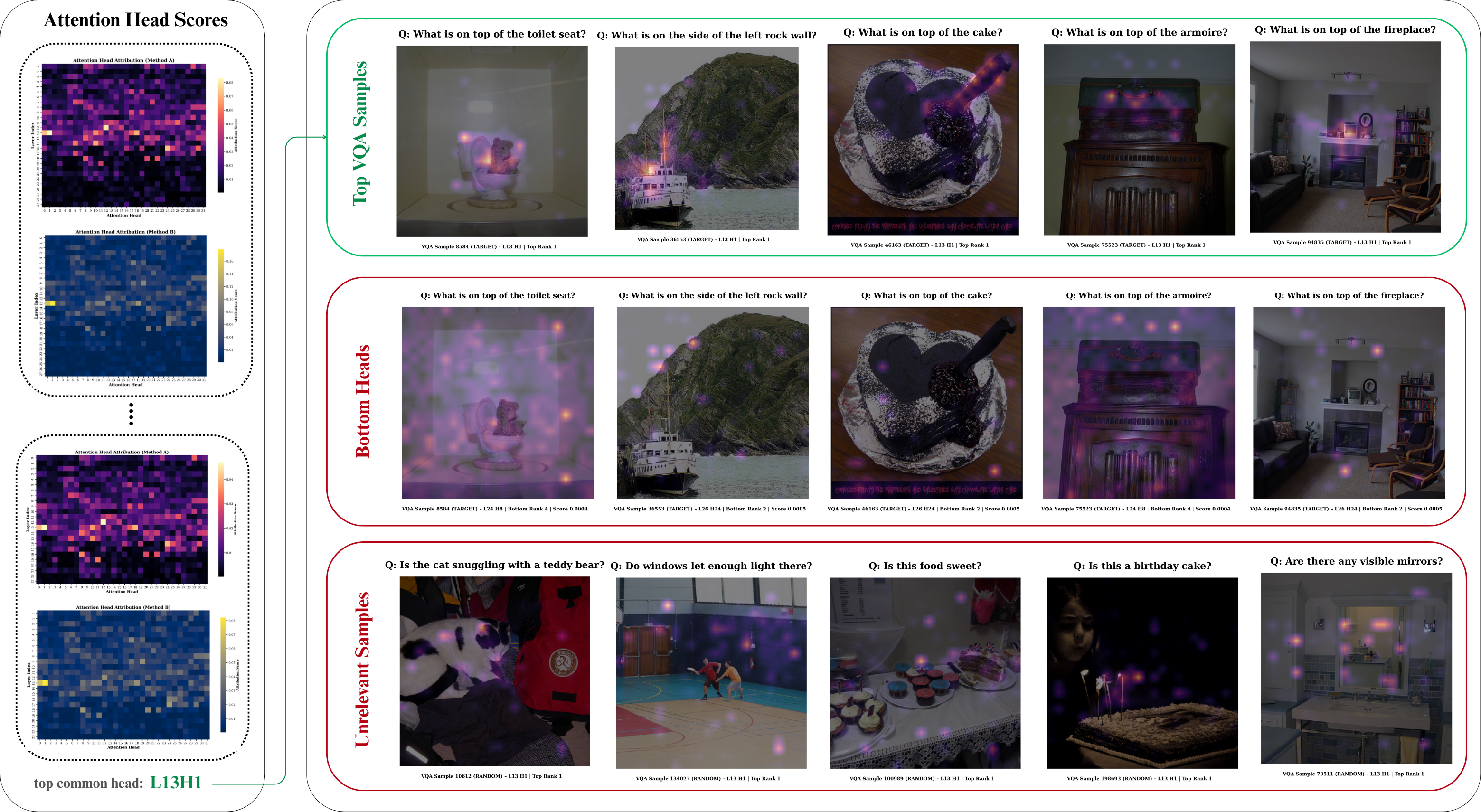}
  \caption{Attribution patching across related spatial features. 
  Top: recurring top-scoring head (L13H1) localizes to relevant regions in queries about “on top of” relations. 
  Middle: bottom-ranked heads on the same samples fail to capture spatial structure. 
  Bottom: unrelated queries confirm that the top head does not spuriously activate.}
  \label{fig:ap-main}
\end{figure*}

To illustrate the effect of attribution patching on individual features, 
Appx.~Fig.~\ref{fig:ap-single-all} provides detailed examples. 
In each case, attribution scores isolate a handful of heads, and qualitative maps confirm 
that high-scoring heads focus on regions consistent with the queried relation (e.g., “on top of,” “behind”), 
whereas low-scoring heads fail to do so. These head-level overlays can also be used to
(i) improve the confidence of automated feature interpretation by coupling sample activations with attention visualizations, and
(ii) examine failure cases by checking whether the top spatial features and heads attend to valid regions in misclassified samples. Interestingly, when we look across multiple related spatial features together, we find that some of the same heads 
recur across related spatial relations. Fig~\ref{fig:ap-main} illustrates this pattern. In the top row, L13H1 attends to semantically relevant regions across queries. As a control, the middle row shows that bottom-ranked heads on the same samples fail to localize meaningfully. The bottom row further confirms that irrelevant queries do not trigger spurious activation.
More generally, these same heads also attend to meaningful regions such as salient objects or attributes under custom prompts 
(Appx.~Fig.~\ref{fig:semantic-heads}), underscoring that attribution patching identifies a set of heads 
that reliably carry spatial–semantic signal. 

% \paragraph{Feature Ablation on VSR.}
% We further test the causal role of spatial features by projecting them out during VSR evaluation. 
% For a decoder vector $v$ (unit norm), we apply $y \leftarrow y - (y v)v$ to the residual writes of attention and MLP outputs, leaving image tokens unchanged. 
% Ablating a recurring spatial feature (e.g., L26F807) on the VSR subset containing its associated spatial cues (e.g., behind, below, beneath, under) reduces accuracy from 75.8\% to 59.3\% ($\Delta=-16.5$ pp) and lowers mean $P(\text{correct})$ from $0.624$ to $0.594$ ($\Delta=-0.029$). 
% Random feature ablations have a negligible effect, confirming that these SAE directions encode functional spatial information used by the model.

\subsection{Ablation Study}

We test whether adapted SAE features are \emph{causally involved} in spatial reasoning by ablating them during inference and measuring performance on VSR~\citep{liu2023vsr}, a dataset of text–image pairs spanning dozens of spatial relations, and on a \textit{Yes/No} subset of VQAv2 (general). Each feature is evaluated on a \emph{relation-specific subset} of VSR constructed from its top-activating samples, so that the ablation directly targets the relation it most strongly encodes. To ablate a target feature $f$ at layer $L$, we orthogonally remove its decoder direction $v$ (unit norm) from the residual stream at \emph{text} token positions, leaving image tokens unchanged:
\[
y \leftarrow y - (y^\top v)\, v .
\]

\paragraph{Evaluation metrics.}
We report: (i) accuracy drop on VSR ($\Delta$VSR Acc; $\downarrow$ is worse), 
(ii) accuracy drop on VQA ($\Delta$VQA Acc), 
(iii) accuracy drop from ablating same-layer random features ($\Delta$Ctrl), 
and (iv) odds ratio under the spatial distribution shift (VSR OR; $\uparrow$ is better). All runs use identical cached indices, and results are averaged over seeds.

\begin{table}[h]
\centering
\small
\setlength{\tabcolsep}{3pt}
\renewcommand{\arraystretch}{1.05}
\resizebox{\columnwidth}{!}{%
\begin{tabular}{
c c
S[table-format=-2.2]
S[table-format=-1.2]
S[table-format=-1.2]
p{2.3cm}
S[table-format=1.2]
}
\toprule
Layer & Feature & {$\Delta$VSR Acc} & {$\Delta$VQA Acc} & {$\Delta$Ctrl} & VSR Relation & {VSR OR} \\
\midrule
7  & 15870 & -15.54 & -0.10 & -0.88 & above & 4.32 \\
11 & 27061 & -12.77 & -0.40 &  0.00 & across from & 8.03 \\
9  & 15404 & -11.19 & -0.80 &  1.08 & below & 5.60 \\
14 & 17873 & -10.21 & -0.30 & -1.71 & at the right side of & 7.17 \\
12 & 23874 &  -9.05 & -0.40 & -0.95 & left of & 9.10 \\
18 & 29948 &  -7.98 & -0.30 &  0.00 & beside & 8.36 \\
\bottomrule
\end{tabular}
}
\caption{\textbf{Top ablated SAE features} ranked by VSR accuracy drop.
Large $\Delta$VSR Acc with small $\Delta$VQA Acc indicates spatial specificity; near-zero $\Delta$Ctrl confirms robustness.}
\vspace{-1em}
\label{tab:ablation-top}
\end{table}

\paragraph{Interpretation.}
Ablating the top spatial features lowers VSR accuracy by 9--16 points on average while leaving general VQA nearly unchanged ($\leq$~1 pp), indicating that these directions are functionally used for spatial reasoning rather than general behavior. This shows that probing or switching off a single feature can selectively disable spatial reasoning without harming overall ability. High odds ratios further show selective recruitment under spatial prompts. Random-feature controls yield effects near zero or inconsistent in sign, supporting specificity. Full per-feature results, probability deltas, and seed-wise summaries are reported in Appx.~\ref{app:full-ablation}.

%%%%%%%%%%%%%%%%%%%%%%%%%%%%%%%%%%%%%%%%%%%%%%%%%%%%%%%%%%%%%%%%%%%%%%%%%%%%%%%%%%%%
\section{Limitations}
Our analyses indicate spatial selectivity, but more detailed ablation and steering studies are needed to fully validate causality. Moreover, our experiments are limited to a single model (LLaVA-More with a LLaMA-3.1-8B backbone); applying the method to other backbones and larger corpora will be key to assessing generality.

%%%%%%%%%%%%%%%%%%%%%%%%%%%%%%%%%%%%%%%%%%%%%%%%%%%%%%%%%%%%%%%%%%%%%%%%%%%%%%%%%%%%

\section{Conclusion}
We set out to understand how a pretrained language backbone learns to “see” under multimodal fine-tuning. By extending stage-wise model diffing to the vision–language setting, we isolated vision-preferring features that undergo strong rotations during training, showed that a subset reliably encodes spatial relations, and traced their causal drivers to a small number of mid-layer attention heads. These results show that multimodal adaptation is structured and interpretable as it can be localized, probed, and explained at the feature level. Beyond spatial reasoning, our methodology offers a general framework for uncovering when and where new capabilities emerge in large models, showing that multimodal adaptation follows structured patterns rather than diffuse changes. We view this work as an early step toward a mechanistic science of multimodal training, where models can be interpreted both in terms of their outputs and the internal features that support them.

%%%%%%%%%%%%%%%%%%%%%%%%%%%%%%%%%%%%%%%%%%%%%%%%%%%%%%%%%%%%%%%%%%%%%%%%%%%%%%%%%%%%

% \section*{References}
%%%%%%%%%%%%%%%%%%%%%%%%%%%%%%%%%%%%%%%%%%%%%%%%%%%%%%%%%%%%%%%%%%%%%%%%%%%%%%

% \printbibliography
% \nocitepp{*}
% \bibliography{references}
% \bibliographystyle{iclr2026_conference}

%%%%%%%%%%%%%%%%%%%%%%%%%%%%%%%%%%%%%%%%%%%%%%%%%%%%%%%%%%%%%%%%%%%%%%%%%%%%%%
\bibliographystyle{icml2026}
\bibliography{references}

@inproceedings{radford2021learning,
  title={Learning transferable visual models from natural language supervision},
  author={Radford, Alec and Kim, Jong Wook and Hallacy, Chris and Ramesh, Aditya and Goh, Gabriel and Agarwal, Sandhini and Sastry, Girish and Askell, Amanda and Mishkin, Pamela and Clark, Jack and others},
  booktitle={International conference on machine learning},
  pages={8748--8763},
  year={2021},
  organization={PmLR}
}

@article{cocchi2025llava,
  title={LLaVA-MORE: A Comparative Study of LLMs and Visual Backbones for Enhanced Visual Instruction Tuning},
  author={Cocchi, Federico and Moratelli, Nicholas and Caffagni, Davide and Sarto, Sara and Baraldi, Lorenzo and Cornia, Marcella and Cucchiara, Rita},
  journal={arXiv preprint arXiv:2503.15621},
  year={2025}
}

@article{liu2023visual,
  title={Visual instruction tuning},
  author={Liu, Haotian and Li, Chunyuan and Wu, Qingyang and Lee, Yong Jae},
  journal={Advances in neural information processing systems},
  volume={36},
  pages={34892--34916},
  year={2023}
}

@inproceedings{liu2024improved,
  title={Improved baselines with visual instruction tuning},
  author={Liu, Haotian and Li, Chunyuan and Li, Yuheng and Lee, Yong Jae},
  booktitle={Proceedings of the IEEE/CVF conference on computer vision and pattern recognition},
  pages={26296--26306},
  year={2024}
}

@article{grattafiori2024llama,
  title={The llama 3 herd of models},
  author={Grattafiori, Aaron and Dubey, Abhimanyu and Jauhri, Abhinav and Pandey, Abhinav and Kadian, Abhishek and Al-Dahle, Ahmad and Letman, Aiesha and Mathur, Akhil and Schelten, Alan and Vaughan, Alex and others},
  journal={arXiv preprint arXiv:2407.21783},
  year={2024}
}

@article{bricken2023monosemanticity,
   title={Towards Monosemanticity: Decomposing Language Models With Dictionary Learning},
   author={Bricken, Trenton and Templeton, Adly and Batson, Joshua and Chen, Brian and Jermyn, Adam and Conerly, Tom and Turner, Nick and Anil, Cem and Denison, Carson and Askell, Amanda and Lasenby, Robert and Wu, Yifan and Kravec, Shauna and Schiefer, Nicholas and Maxwell, Tim and Joseph, Nicholas and Hatfield-Dodds, Zac and Tamkin, Alex and Nguyen, Karina and McLean, Brayden and Burke, Josiah E and Hume, Tristan and Carter, Shan and Henighan, Tom and Olah, Christopher},
   year={2023},
   journal={Transformer Circuits Thread},
   note={https://transformer-circuits.pub/2023/monosemantic-features/index.html}
}

@article{cunningham2023sparse,
  title={Sparse autoencoders find highly interpretable features in language models},
  author={Cunningham, Hoagy and Ewart, Aidan and Riggs, Logan and Huben, Robert and Sharkey, Lee},
  journal={arXiv preprint arXiv:2309.08600},
  year={2023}
}

@article{gao2024scaling,
  title={Scaling and evaluating sparse autoencoders},
  author={Gao, Leo and la Tour, Tom Dupr{\'e} and Tillman, Henk and Goh, Gabriel and Troll, Rajan and Radford, Alec and Sutskever, Ilya and Leike, Jan and Wu, Jeffrey},
  journal={arXiv preprint arXiv:2406.04093},
  year={2024}
}

@article{templeton2024scaling,
       title={Scaling Monosemanticity: Extracting Interpretable Features from Claude 3 Sonnet},
       author={Templeton, Adly and Conerly, Tom and Marcus, Jonathan and Lindsey, Jack and Bricken, Trenton and Chen, Brian and Pearce, Adam and Citro, Craig and Ameisen, Emmanuel and Jones, Andy and Cunningham, Hoagy and Turner, Nicholas L and McDougall, Callum and MacDiarmid, Monte and Freeman, C. Daniel and Sumers, Theodore R. and Rees, Edward and Batson, Joshua and Jermyn, Adam and Carter, Shan and Olah, Chris and Henighan, Tom},
       year={2024},
       journal={Transformer Circuits Thread},
       url={https://transformer-circuits.pub/2024/scaling-monosemanticity/index.html}
}

@article{minder2025robustly,
  title={Robustly identifying concepts introduced during chat fine-tuning using crosscoders},
  author={Minder, Julian and Dumas, Cl{\'e}ment and Juang, Caden and Chugtai, Bilal and Nanda, Neel},
  journal={arXiv preprint arXiv:2504.02922},
  year={2025}
}

@article{Bricken2024StageWiseModelDiffing,
  title        = {Stage-Wise Model Diffing},
  author       = {Trenton Bricken and Siddharth Mishra-Sharma and Jonathan Marcus and Adam Jermyn and Christopher Olah and Kelley Rivoire and Thomas Henighan},
  year         = {2024},
  note         = {\url{https://transformer-circuits.pub/2024/model-diffing/index.html}}
}

@misc{olah2015visualizing,
  title        = {Visualizing Representations: Deep Learning and Human Beings},
  author       = {Olah, Christopher},
  year         = {2015},
  howpublished = {\url{https://colah.github.io/posts/2015-01-Visualizing-Representations/}},
  note         = {Accessed: 2025-08-23}
}

@article{erhan2010,
author = {Erhan, Dumitru and Bengio, Yoshua and Courville, Aaron and Manzagol, Pierre-Antoine and Vincent, Pascal and Bengio, Samy},
title = {Why Does Unsupervised Pre-training Help Deep Learning?},
year = {2010},
issue_date = {3/1/2010},
publisher = {JMLR.org},
volume = {11},
issn = {1532-4435},
month = mar,
pages = {625–660},
numpages = {36}
}

@inproceedings{lenc2015understanding,
  title={Understanding image representations by measuring their equivariance and equivalence},
  author={Lenc, Karel and Vedaldi, Andrea},
  booktitle={Proceedings of the IEEE conference on computer vision and pattern recognition},
  pages={991--999},
  year={2015}
}

@article{bansal2021revisiting,
  title={Revisiting model stitching to compare neural representations},
  author={Bansal, Yamini and Nakkiran, Preetum and Barak, Boaz},
  journal={Advances in neural information processing systems},
  volume={34},
  pages={225--236},
  year={2021}
}

@inproceedings{kornblith2019similarity,
  title={Similarity of neural network representations revisited},
  author={Kornblith, Simon and Norouzi, Mohammad and Lee, Honglak and Hinton, Geoffrey},
  booktitle={International conference on machine learning},
  pages={3519--3529},
  year={2019},
  organization={PMlR}
}

@article{barannikov2021representation,
  title={Representation topology divergence: A method for comparing neural network representations},
  author={Barannikov, Serguei and Trofimov, Ilya and Balabin, Nikita and Burnaev, Evgeny},
  journal={arXiv preprint arXiv:2201.00058},
  year={2021}
}

@article{li2015convergent,
  title={Convergent learning: Do different neural networks learn the same representations?},
  author={Li, Yixuan and Yosinski, Jason and Clune, Jeff and Lipson, Hod and Hopcroft, John},
  journal={arXiv preprint arXiv:1511.07543},
  year={2015}
}

@article{olah2020zoom,
  author = {Olah, Chris and Cammarata, Nick and Schubert, Ludwig and Goh, Gabriel and Petrov, Michael and Carter, Shan},
  title = {Zoom In: An Introduction to Circuits},
  journal = {Distill},
  year = {2020},
  note = {https://distill.pub/2020/circuits/zoom-in},
  doi = {10.23915/distill.00024.001}
}

@article{khayatan2025analyzing,
  title={Analyzing Fine-tuning Representation Shift for Multimodal LLMs Steering alignment},
  author={Khayatan, Pegah and Shukor, Mustafa and Parekh, Jayneel and Cord, Matthieu},
  journal={arXiv preprint arXiv:2501.03012},
  year={2025}
}

@article{venhoff2025visual,
  title={How Visual Representations Map to Language Feature Space in Multimodal LLMs},
  author={Venhoff, Constantin and Khakzar, Ashkan and Joseph, Sonia and Torr, Philip and Nanda, Neel},
  journal={arXiv preprint arXiv:2506.11976},
  year={2025}
}

@unpublished{kissane2024saes,
  author       = {Kissane, Connor and Krzyzanowski, Robert and Conmy, Arthur and Nanda, Neel},
  title        = {SAEs (usually) Transfer Between Base and Chat Models},
  note         = {AI Alignment Forum post},
      year         = {2024},
  month        = jul # "~18",
  howpublished = {Interim report on AI Alignment Forum}
}

@article{stan2024lvlm,
  title={LVLM-Interpret: an interpretability tool for large vision-language models},
  author={Stan, Gabriela Ben Melech and Aflalo, Estelle and Rohekar, Raanan Yehezkel and Bhiwandiwalla, Anahita and Tseng, Shao-Yen and Olson, Matthew Lyle and Gurwicz, Yaniv and Wu, Chenfei and Duan, Nan and Lal, Vasudev},
  journal={arXiv preprint arXiv:2404.03118},
  year={2024}
}

@article{basu2024understanding,
  title={Understanding information storage and transfer in multi-modal large language models},
  author={Basu, Samyadeep and Grayson, Martin and Morrison, Cecily and Nushi, Besmira and Feizi, Soheil and Massiceti, Daniela},
  journal={Advances in Neural Information Processing Systems},
  volume={37},
  pages={7400--7426},
  year={2024}
}

@inproceedings{tong2024eyes,
  title={Eyes wide shut? exploring the visual shortcomings of multimodal llms},
  author={Tong, Shengbang and Liu, Zhuang and Zhai, Yuexiang and Ma, Yi and LeCun, Yann and Xie, Saining},
  booktitle={Proceedings of the IEEE/CVF Conference on Computer Vision and Pattern Recognition},
  pages={9568--9578},
  year={2024}
}

@article{gandelsman2023interpreting,
  title={Interpreting clip's image representation via text-based decomposition},
  author={Gandelsman, Yossi and Efros, Alexei A and Steinhardt, Jacob},
  journal={arXiv preprint arXiv:2310.05916},
  year={2023}
}

@article{chen2023interpreting,
  title={Interpreting and controlling vision foundation models via text explanations},
  author={Chen, Haozhe and Yang, Junfeng and Vondrick, Carl and Mao, Chengzhi},
  journal={arXiv preprint arXiv:2310.10591},
  year={2023}
}

@inproceedings{schwettmann2023multimodal,
  title={Multimodal neurons in pretrained text-only transformers},
  author={Schwettmann, Sarah and Chowdhury, Neil and Klein, Samuel and Bau, David and Torralba, Antonio},
  booktitle={Proceedings of the IEEE/CVF International Conference on Computer Vision},
  pages={2862--2867},
  year={2023}
}

@article{jiang2024interpreting,
  title={Interpreting and editing vision-language representations to mitigate hallucinations},
  author={Jiang, Nick and Kachinthaya, Anish and Petryk, Suzie and Gandelsman, Yossi},
  journal={arXiv preprint arXiv:2410.02762},
  year={2024}
}

@inproceedings{palit2023towards,
  title={Towards vision-language mechanistic interpretability: A causal tracing tool for blip},
  author={Palit, Vedant and Pandey, Rohan and Arora, Aryaman and Liang, Paul Pu},
  booktitle={Proceedings of the IEEE/CVF International Conference on Computer Vision},
  pages={2856--2861},
  year={2023}
}

@article{neo2024towards,
  title={Towards interpreting visual information processing in vision-language models},
  author={Neo, Clement and Ong, Luke and Torr, Philip and Geva, Mor and Krueger, David and Barez, Fazl},
  journal={arXiv preprint arXiv:2410.07149},
  year={2024}
}

@inproceedings{venhoff2025toolate,
  title        = {Too Late to Recall: The Two-Hop Problem in Multimodal Knowledge Retrieval},
  author       = {Constantin Venhoff and Ashkan Khakzar and Sonia Joseph and Philip Torr and Neel Nanda},
  booktitle    = {Mechanistic Interpretability for Vision (Non-proceedings Track), CVPR 2025},
  year         = {2025},
  url          = {https://openreview.net/forum?id=VUhRdZp8ke},
}

@inproceedings{venhoff2025twohop,
  title     = {Too Late to Recall: The Two-Hop Problem in Multimodal Knowledge Retrieval},
  author    = {Venhoff, Constantin and Khakzar, Ashkan and Joseph, Sonia and Torr, Philip and Nanda, Neel},
  booktitle = {CVPR 2025 Workshop on Mechanistic Interpretability of Vision (MIV)},
  year      = {2025},
  note      = {Non-proceedings Track Poster},
}

@misc{nanda2023attribution,
  author       = {Neel Nanda},
  title        = {Attribution Patching: Activation Patching at Industrial Scale},
  year         = {2023},
  howpublished = {\url{https://www.neelnanda.io/mechanistic-interpretability}},
  note         = {Accessed: 2025-08-23}
}

@inproceedings{zhang2024activationpatching,
  title     = {Towards Best Practices of Activation Patching in Language Models: Metrics and Methods},
  author    = {Zhang, Fred and Nanda, Neel},
  booktitle = {International Conference on Learning Representations (ICLR)},
  year      = {2024},
  url       = {https://doi.org/10.48550/arXiv.2309.16042},
  note      = {arXiv:2309.16042}
}

@article{he2024llama,
  title={Llama scope: Extracting millions of features from llama-3.1-8b with sparse autoencoders},
  author={He, Zhengfu and Shu, Wentao and Ge, Xuyang and Chen, Lingjie and Wang, Junxuan and Zhou, Yunhua and Liu, Frances and Guo, Qipeng and Huang, Xuanjing and Wu, Zuxuan and others},
  journal={arXiv preprint arXiv:2410.20526},
  year={2024}
}

@article{hubinger2024sleeper,
  title={Sleeper agents: Training deceptive llms that persist through safety training},
  author={Hubinger, Evan and Denison, Carson and Mu, Jesse and Lambert, Mike and Tong, Meg and MacDiarmid, Monte and Lanham, Tamera and Ziegler, Daniel M and Maxwell, Tim and Cheng, Newton and others},
  journal={arXiv preprint arXiv:2401.05566},
  year={2024}
}

@misc{mistral2024pixtral,
  author       = {Mistral AI},
  title        = {Pixtral 12B: A New Frontier in Image and Text Understanding},
  howpublished = {\url{https://mistral.ai/news/pixtral-12b/}},
  year         = {2024},
  month        = {September},
  day          = {17},
  note         = {Accessed: 2024-12-21}
}

@article{zhang2024improve,
  title={Improve vision language model chain-of-thought reasoning},
  author={Zhang, Ruohong and Zhang, Bowen and Li, Yanghao and Zhang, Haotian and Sun, Zhiqing and Gan, Zhe and Yang, Yinfei and Pang, Ruoming and Yang, Yiming},
  journal={arXiv preprint arXiv:2410.16198},
  year={2024}
}

@article{dong2024insight,
  title={Insight-V: Exploring Long-Chain Visual Reasoning with Multimodal Large Language Models},
  author={Dong, Yuhao and Liu, Zuyan and Sun, Hai-Long and Yang, Jingkang and Hu, Winston and Rao, Yongming and Liu, Ziwei},
  journal={arXiv preprint arXiv:2411.14432},
  year={2024}
}

@inproceedings{goyal2017making,
  title={Making the v in vqa matter: Elevating the role of image understanding in visual question answering},
  author={Goyal, Yash and Khot, Tejas and Summers-Stay, Douglas and Batra, Dhruv and Parikh, Devi},
  booktitle={Proceedings of the IEEE conference on computer vision and pattern recognition},
  pages={6904--6913},
  year={2017}
}

@misc{gpt-4o-mini,
  title = {GPT-4O-Mini: Advancing Cost-Efficient Intelligence},
  author = {OpenAI},
  year = {2024},
  howpublished = {\url{https://openai.com/index/gpt-4o-mini-advancing-cost-efficient-intelligence/}},
  note = {Accessed: 2024-12-21}
}

@article{li2024llava,
  title={Llava-next-interleave: Tackling multi-image, video, and 3d in large multimodal models},
  author={Li, Feng and Zhang, Renrui and Zhang, Hao and Zhang, Yuanhan and Li, Bo and Li, Wei and Ma, Zejun and Li, Chunyuan},
  journal={arXiv preprint arXiv:2407.07895},
  year={2024}
}

@inproceedings{Dong2024ProgressiveMR,
  title={Progressive Multimodal Reasoning via Active Retrieval},
  author={Guanting Dong and Chenghao Zhang and Mengjie Deng and Yutao Zhu and Zhicheng Dou and Ji-Rong Wen},
  year={2024},
  url={https://api.semanticscholar.org/CorpusID:274859457}
}

@article{xu2024llava,
  title={LLaVA-o1: Let Vision Language Models Reason Step-by-Step},
  author={Xu, Guowei and Jin, Peng and Hao, Li and Song, Yibing and Sun, Lichao and Yuan, Li},
  journal={arXiv preprint arXiv:2411.10440},
  year={2024}
}

@article{liu2023vsr,
  title={Visual Spatial Reasoning},
  author={Fangyu Liu and Guy Edward Toh Emerson and Nigel Collier},
  journal={Transactions of the Association for Computational Linguistics},
  year={2023},
}

@article{kantamneni2025saeprobes,
  title={Are Sparse Autoencoders Useful? A Case Study in Sparse Probing},
  author={Kantamneni, Subhash and Engels, Joshua and Rajamanoharan, Senthooran and Tegmark, Max and Nanda, Neel},
  journal={arXiv preprint arXiv:2502.16681},
  year={2025},
}

@article{minder2025diff,
  title={What We Learned Trying to Diff Base and Chat Models (And Why It Matters)},
  author={Minder, Julian and Dumas, Clément and Nanda, Neel},
  journal={LessWrong},
  year={2025}
}

@article{lindsey2024crosscoders,
  title={Sparse Crosscoders for Cross-Layer Features and Model Diffing},
  author={Lindsey, Jack and Tran, Anh Tuan and Nanda, Neel and Bricken, Trenton and Jermyn, Adam and Rivoire, Kelley and Olah, Chris},
  journal={Transformer Circuits Thread},
  year={2024},
  url={https://transformer-circuits.pub/2024/crosscoders/index.html}
}

\clearpage
\newpage

\appendix

\section{Appendix}
\subsection{Geometry divergence: decoder cosine trends}
\label{app:rotations}

\begin{figure*}[t]
  \centering
  \includegraphics[width=0.8\linewidth]{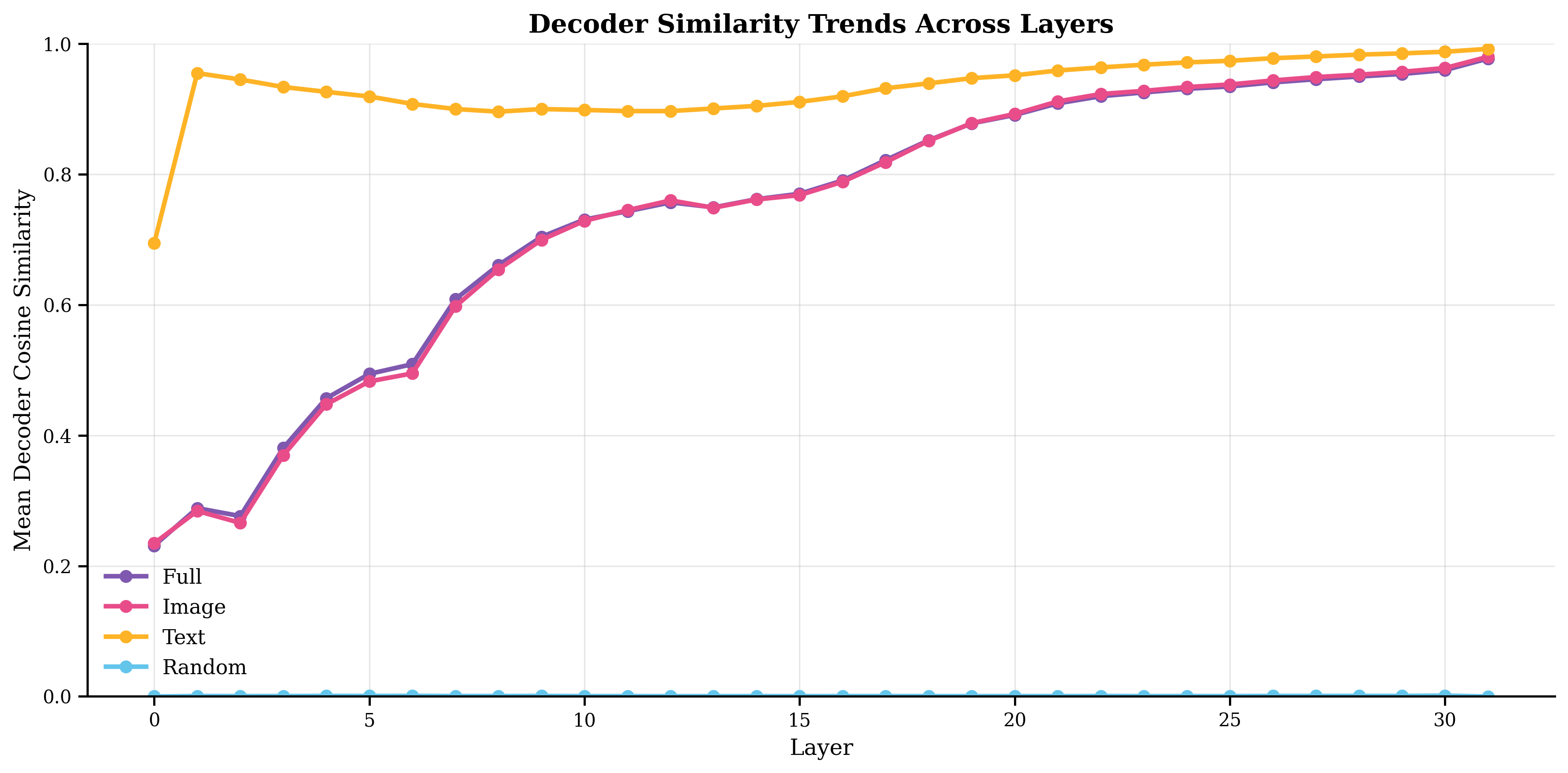}
  \caption{\textbf{Decoder cosine similarity vs.\ layer (LLM SAE vs.\ VLM SAE).}
  Text-only stays highly aligned across layers; image-only and full-sequence rotate in shallow layers and align later; random remains near zero. Higher cosine indicates closer alignment of SAE decoder directions.}
  \label{fig:decoder-cos-trends}
\end{figure*}

To quantify how SAE feature geometry shifts across training regimes, we track cosine similarity between decoder directions from SAEs trained on different input types. Fig~\ref{fig:decoder-cos-trends} shows that text-only SAEs remain closely aligned across layers, while image-only and full-sequence SAEs diverge in early layers before realigning deeper in the model. Randomly initialized SAEs stay largely uncorrelated, confirming the stability of the observed trends.

\subsection{Per-layer FVU trajectories}
\label{app:fvu-layerwise}

\begin{figure*}[t]
\centering
\includegraphics[width=\linewidth]{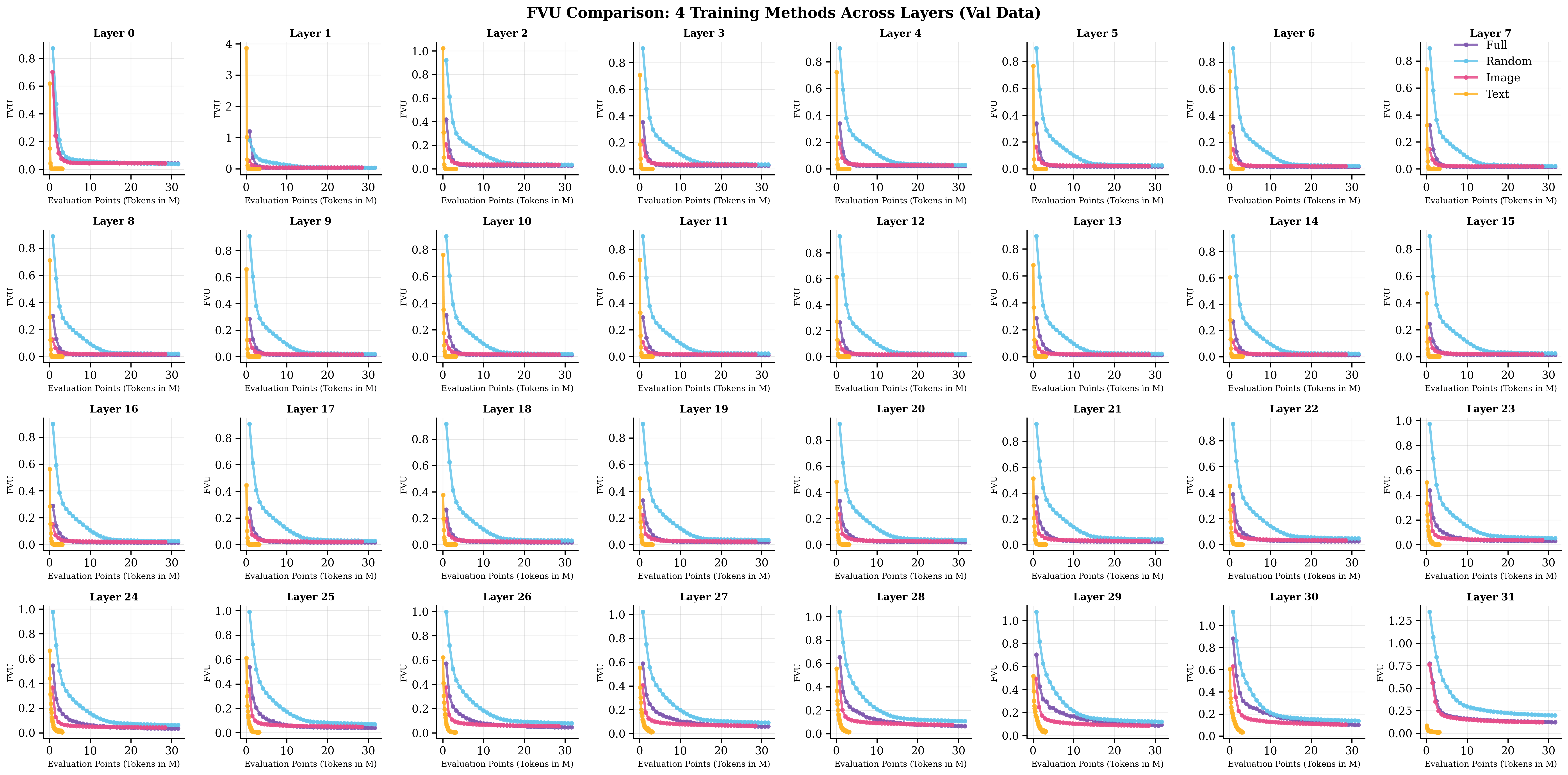}
\caption{\textbf{Per-layer FVU across regimes.} 
Each panel shows the convergence of SAEs trained with different masking regimes for a specific layer. 
Text-only SAEs begin with slightly higher error in the shallowest layers but adapt almost immediately to near-zero reconstruction. 
Image-only and full-sequence SAEs converge more slowly and plateau at higher error, while random initialization performs worst throughout. 
This confirms that projector-driven spans remain off-distribution in early layers and only align with the LLM basis in later layers.}
\label{fig:fvu-layerwise}
\end{figure*}

Fig.~\ref{fig:fvu-layerwise} summarizes per-layer FVU convergence for each masking regime.

\subsection{Per-Layer Statistics}
\label{app:layer-stats}

\begin{figure*}[t]
  \centering
  \begin{subfigure}[h]{0.48\linewidth}
    \centering
    \includegraphics[width=\linewidth]{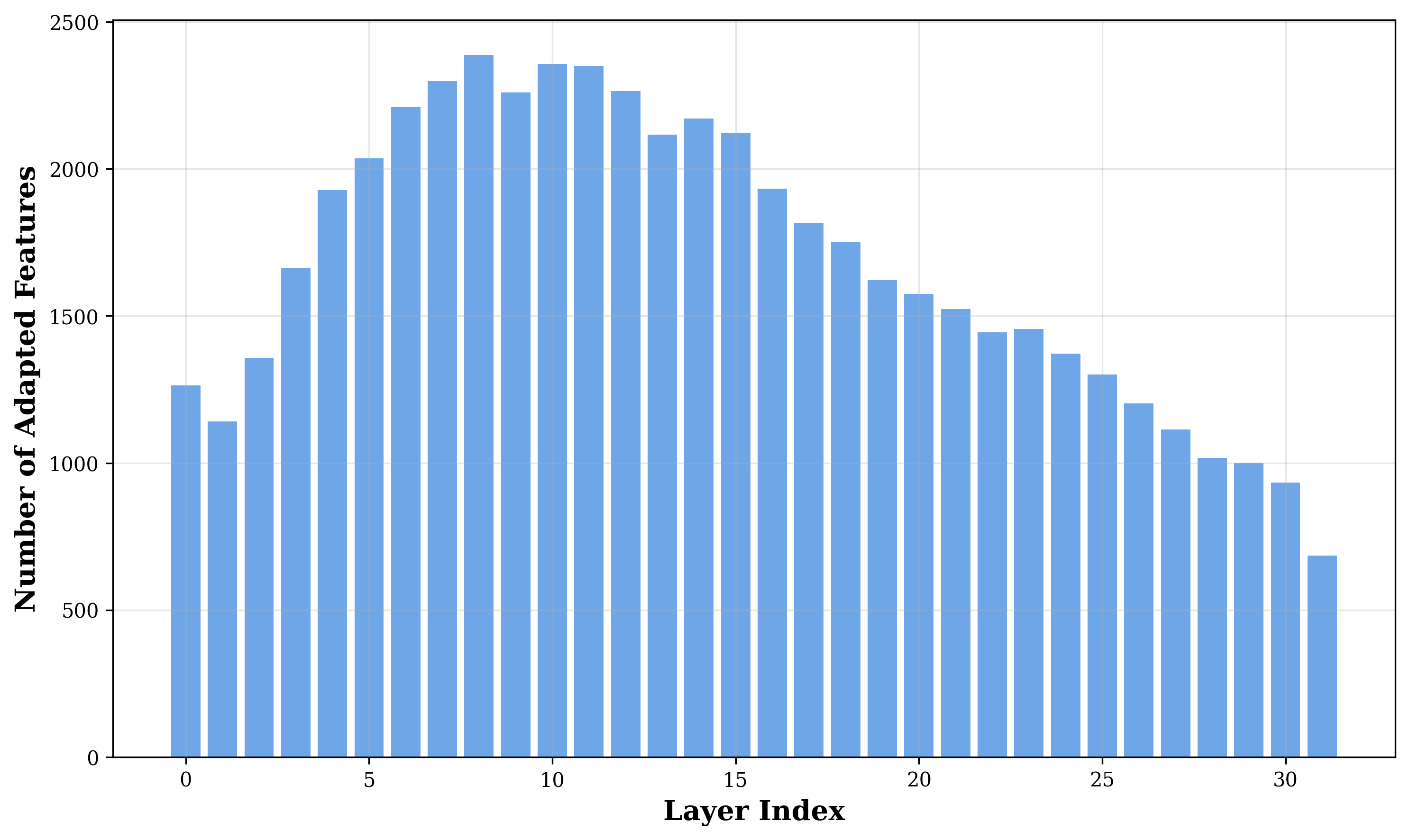}
    \caption{\textbf{Adapted features per layer.} Most concentrate in mid layers, tapering in deeper blocks.}
    \label{fig:suspects-per-layer}
  \end{subfigure}
  \hfill
  \begin{subfigure}[h]{0.48\linewidth}
    \centering
    \includegraphics[width=\linewidth]{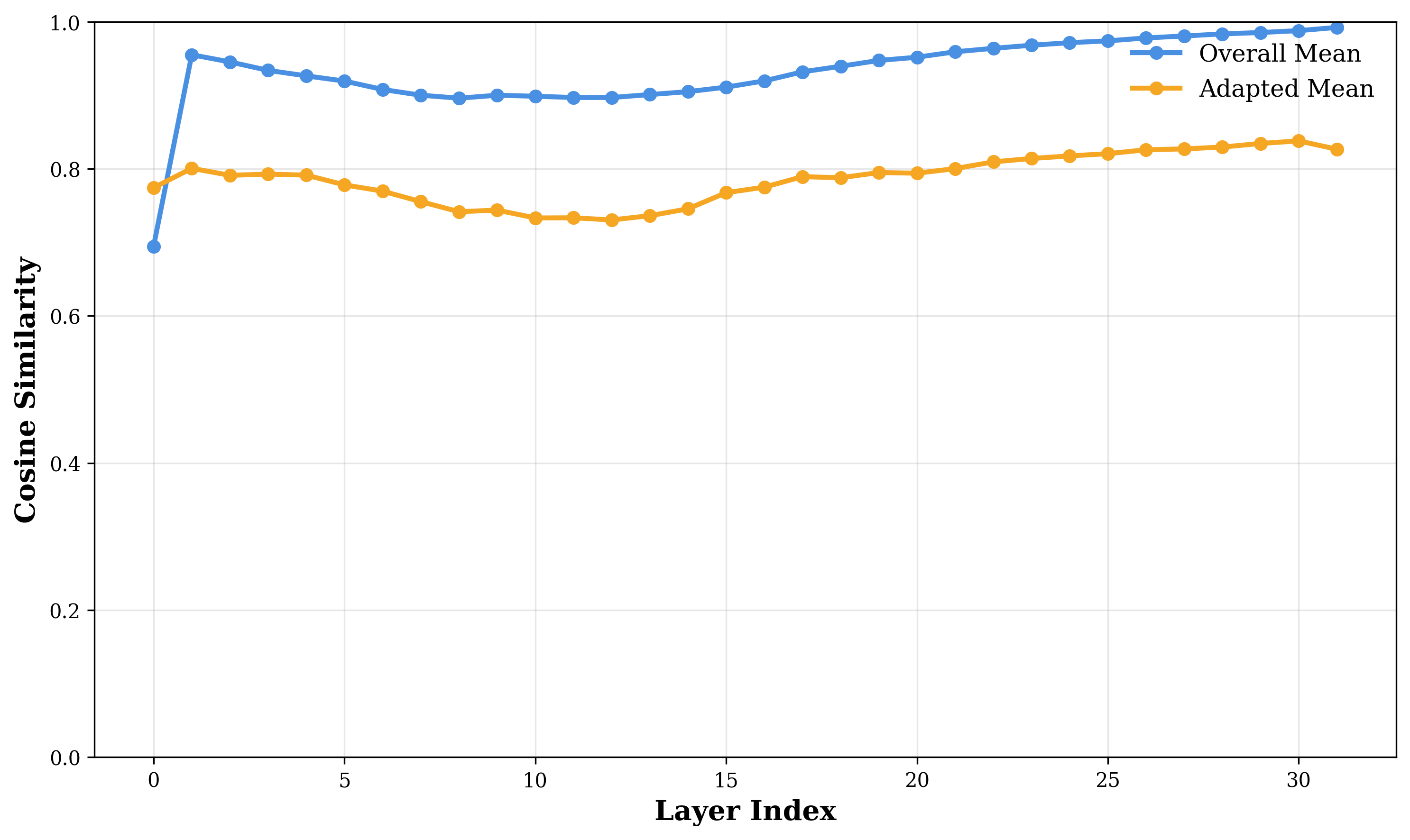}
    \caption{\textbf{Decoder cosine by layer.} Adapted features remain less aligned to the base dictionary than the overall pool.}
    \label{fig:cosine-overall-vs-suspect}
  \end{subfigure}
  \vspace{-0.6em}
  \caption{\textbf{Per-layer statistics of adapted features.} 
  (a) Distribution of adapted feature counts across depth.
  (b) Mean decoder cosine similarity for adapted features vs.\ the overall pool.}
  \label{fig:appendix-layer-stats}
\end{figure*}

Fig~\ref{fig:appendix-layer-stats} shows that adapted features cluster in
mid layers and taper in deeper blocks. Their decoder directions remain less
aligned to the base dictionary than the overall pool, confirming stronger
rotations under multimodal fine-tuning.

\subsection{Threshold Sweep for Feature Selection}
\label{app:threshold-sweep}

\begin{figure*}[t]
    \centering
    \includegraphics[width=\linewidth]{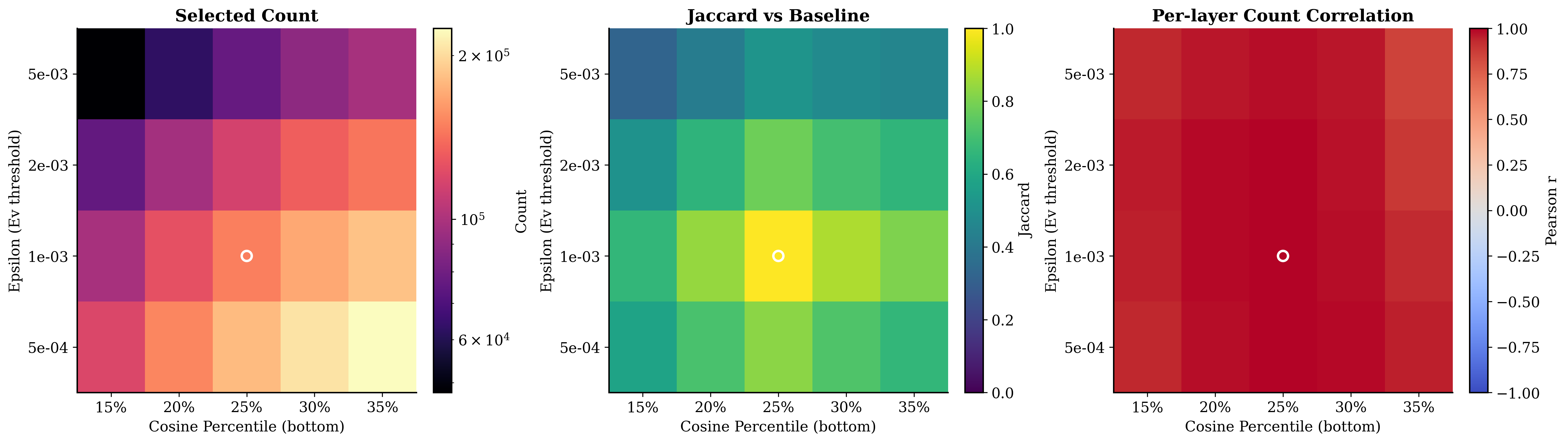}
    \caption{\textbf{Threshold sweep for feature selection.}
    Left: feature counts increase smoothly with more lenient thresholds. 
    Middle: Jaccard overlap with the baseline peaks near the chosen point. 
    Right: per-layer counts remain highly correlated across thresholds. 
    The white circle marks the adopted operating point.}
    \label{fig:threshold-sweep}
\end{figure*}

To ensure that our choice of thresholds is robust, we sweep over the cosine percentile cutoff 
($p_{\cos}$) and visual energy threshold ($\epsilon$). 
Fig.~\ref{fig:threshold-sweep} reports three metrics: (i) total number of selected features, 
(ii) Jaccard overlap with the baseline adapted set, and (iii) per-layer count correlation.
The results show a broad stable region around $\epsilon \approx 10^{-3}$ and $p_{\cos}\approx 25\%$, 
which yields a compact yet consistent set of adapted features. 
We adopt this operating point (white circle) for all downstream analyses.

The visual-energy statistic $E_v$ is computed under a text-only mask, since our SAEs are text-only. 
As a result, most features have $E_v=0$, so requiring $\epsilon>0$ acts as a strong filter. 
When cross-checking with downstream spatial tasks, we find that features with very low $E_v$ rarely contribute meaningfully: they tend to cluster in shallow layers, show low spatial hit rates, and often appear polysemantic on inspection. 
In contrast, those that pass the $\epsilon$ cutoff carry a cleaner visual signal and align more consistently with spatially selective units in downstream evaluations, suggesting that the thresholded set captures genuinely vision-grounded features.

\subsection{Distribution-shift visualizations}
\label{app:dist-shift}

To complement the main-text description of our feature-selection procedure, we include
here the firing-frequency distributions and candidate-feature scatter plots used to
identify spatial units under different prompting conditions.

\begin{figure*}[t]
    \centering
    \begin{subfigure}[t]{0.48\linewidth}
        \centering
        \includegraphics[width=\linewidth]{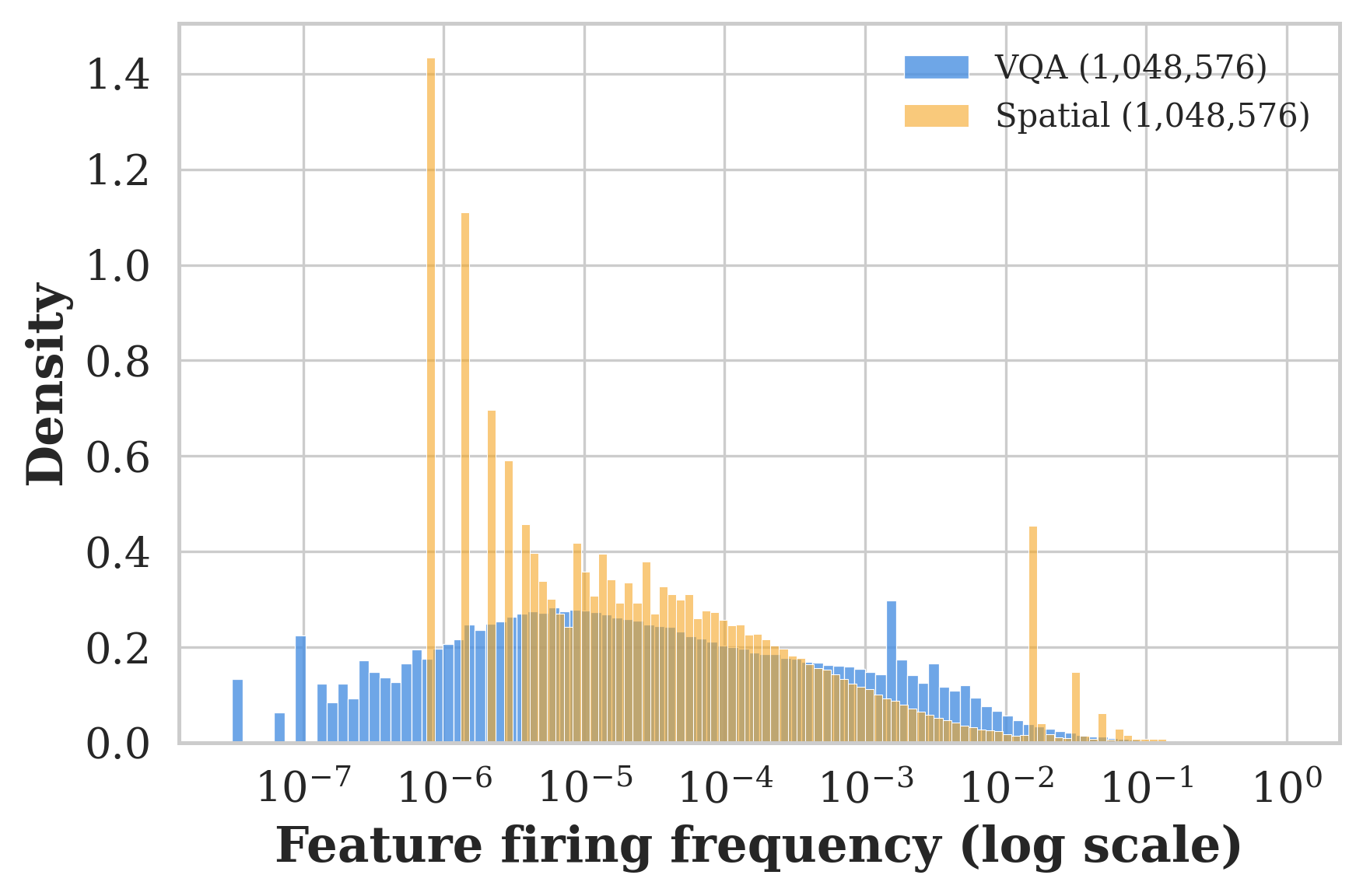}
        \caption{Firing-frequency distributions for $\mathcal{D}_{\text{base}}$
        and the spatial split $\mathcal{D}_{\text{sp}}$.}
        \label{fig:spatial-hist}
    \end{subfigure}
    \hfill
    \begin{subfigure}[t]{0.48\linewidth}
        \centering
        \includegraphics[width=\linewidth]{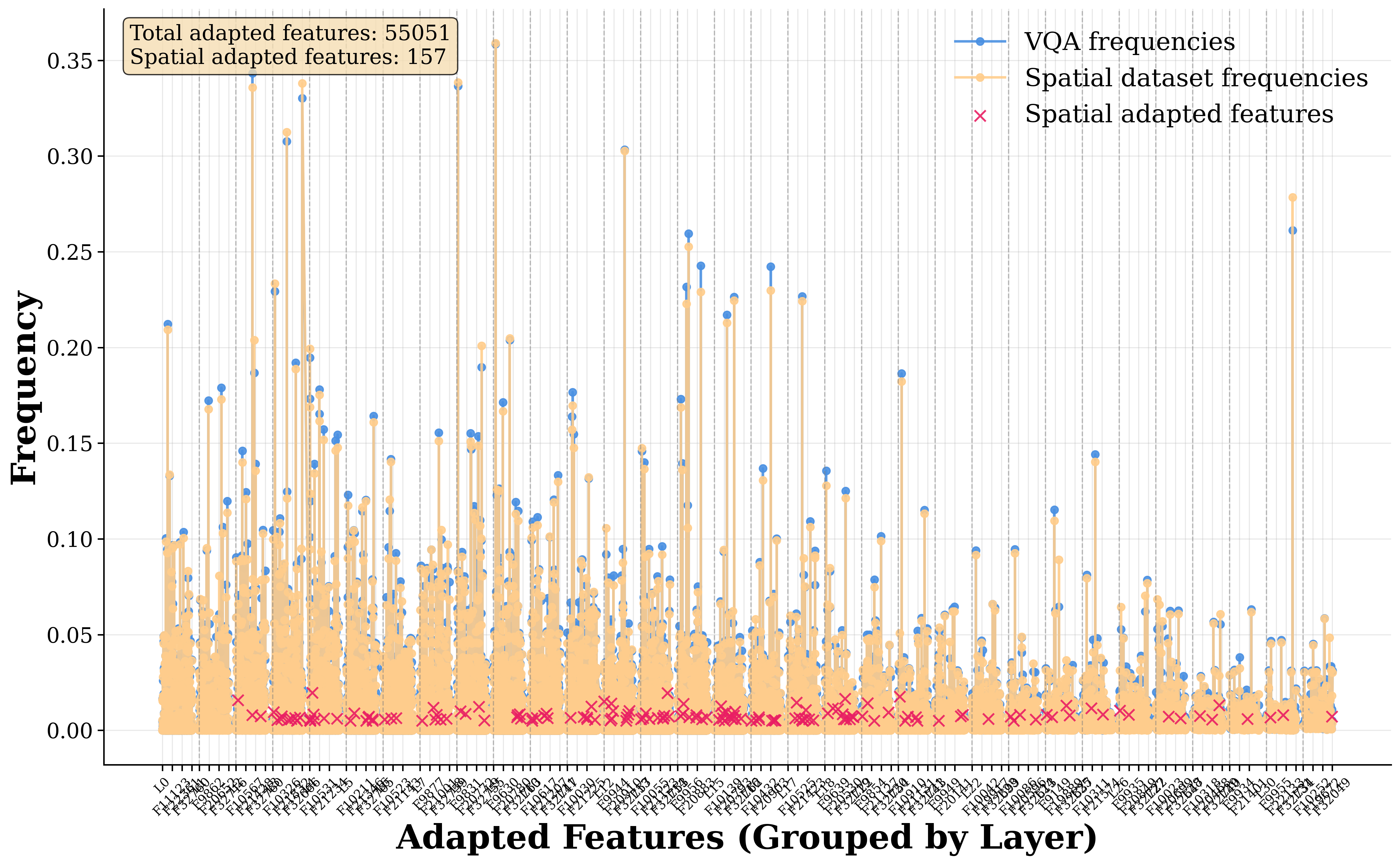}
        \caption{Spatial candidate features under both splits, with
        selected units highlighted.}
        \label{fig:spatial-scatter}
    \end{subfigure}
    \caption{\textbf{Spatial distribution shift.}
    Visualization of feature firing frequencies and candidate selection
    under the spatial vs.\ base splits.}
    \label{fig:spatial-dist-shift}
\end{figure*}

\newpage 
\subsection{OCR feature visualizations}
\label{app:ocr}

\begin{figure*}[t]
  \centering
  \includegraphics[width=0.9\linewidth]{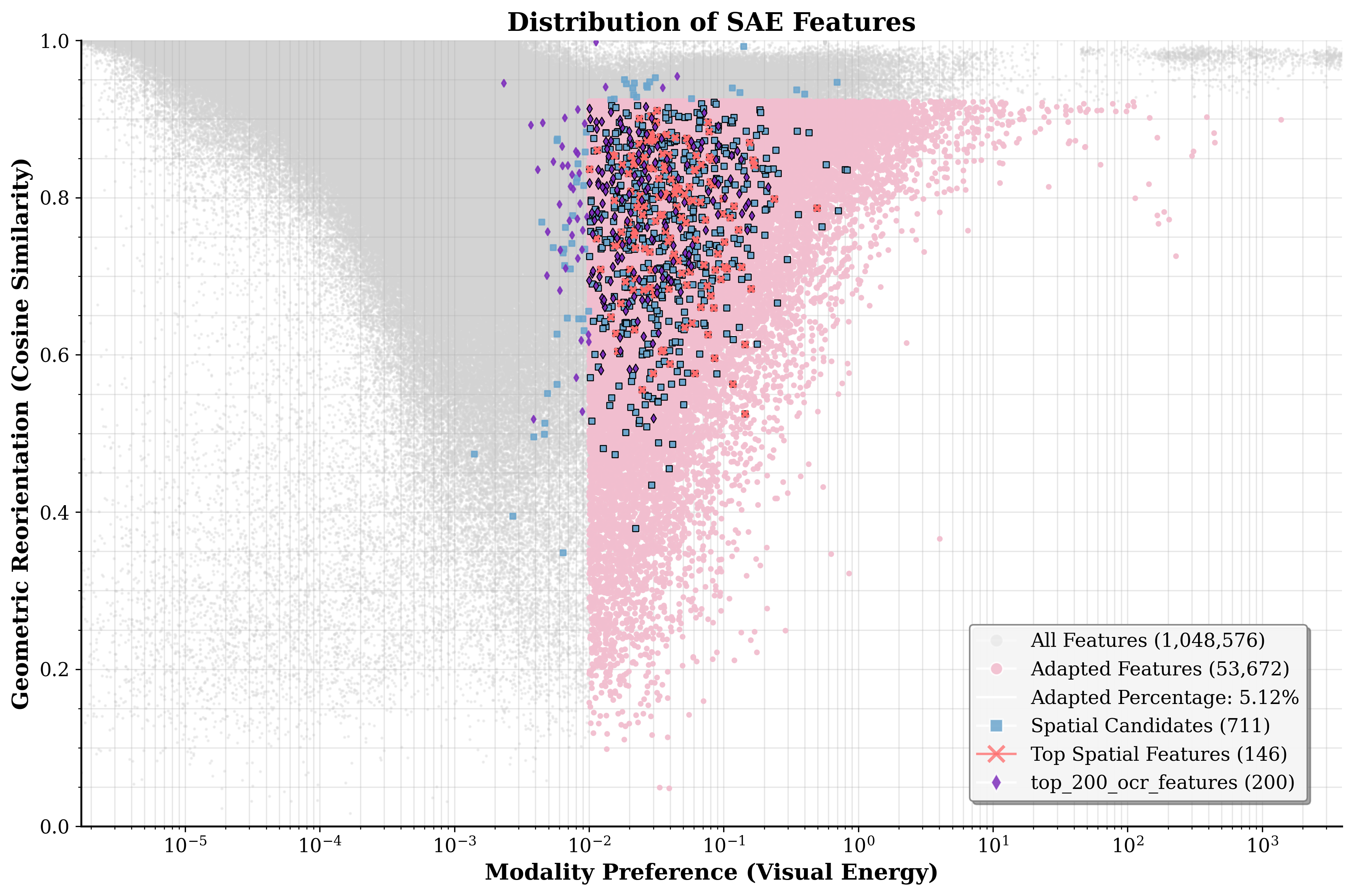}
  \caption{\textbf{Distribution of OCR features.}
  Top OCR candidates (purple) cluster among adapted units (pink), paralleling the
  spatial subset (blue).}
  \label{fig:ocr-scatter}
\end{figure*}

We also apply our distribution-shift procedure to OCR-style prompts (e.g.,
``What does the sign say?''). Fig.~\ref{fig:ocr-scatter} shows that OCR-selective
features cluster within the same adapted region as the spatial subset,
indicating that multimodal fine-tuning concentrates visually grounded capabilities
into a compact envelope of feature space.

\begin{figure*}[t]
  \centering
  \includegraphics[width=0.95\textwidth]{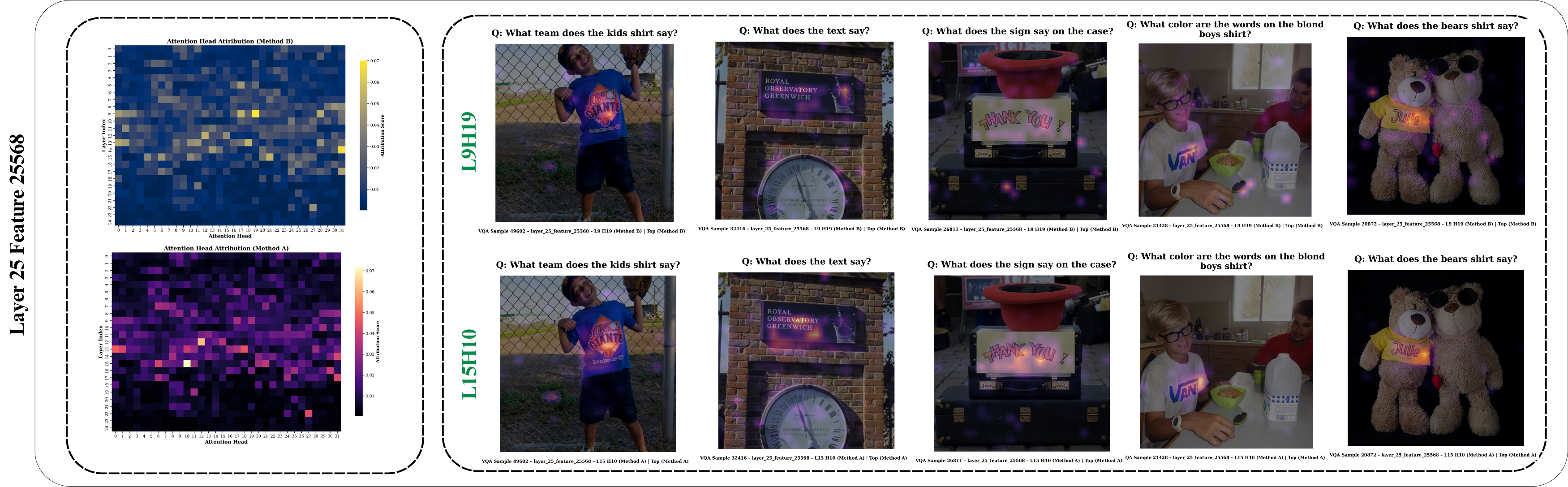}
  \caption{\textbf{Layer 25, Feature 25568.}
  Activates on storefront and clothing text; top heads align to characters.}
  \label{fig:ocr-extra}
\end{figure*}

Qualitative examples confirm that these features reliably activate on embedded
text and that associated heads localise to glyph regions (Fig.
\ref{fig:ocr-extra}), consistent with image-grounded text processing.

\subsection{Additional Auto-Interp Examples}
\label{app:auto-interp}

In the main text (Sec.~\ref{sec:auto-interp}), we showed examples of adapted features using our automated interpretation pipeline. We include two further examples here. In both cases, the top-activating samples agree across VQA and VSR, and the interpretations are consistent and monosemantic.

\begin{figure*}[t]
  \centering
  \vspace{-1.5em}
  \begin{subfigure}[t]{0.9\linewidth}
    \centering
    \includegraphics[width=\linewidth]{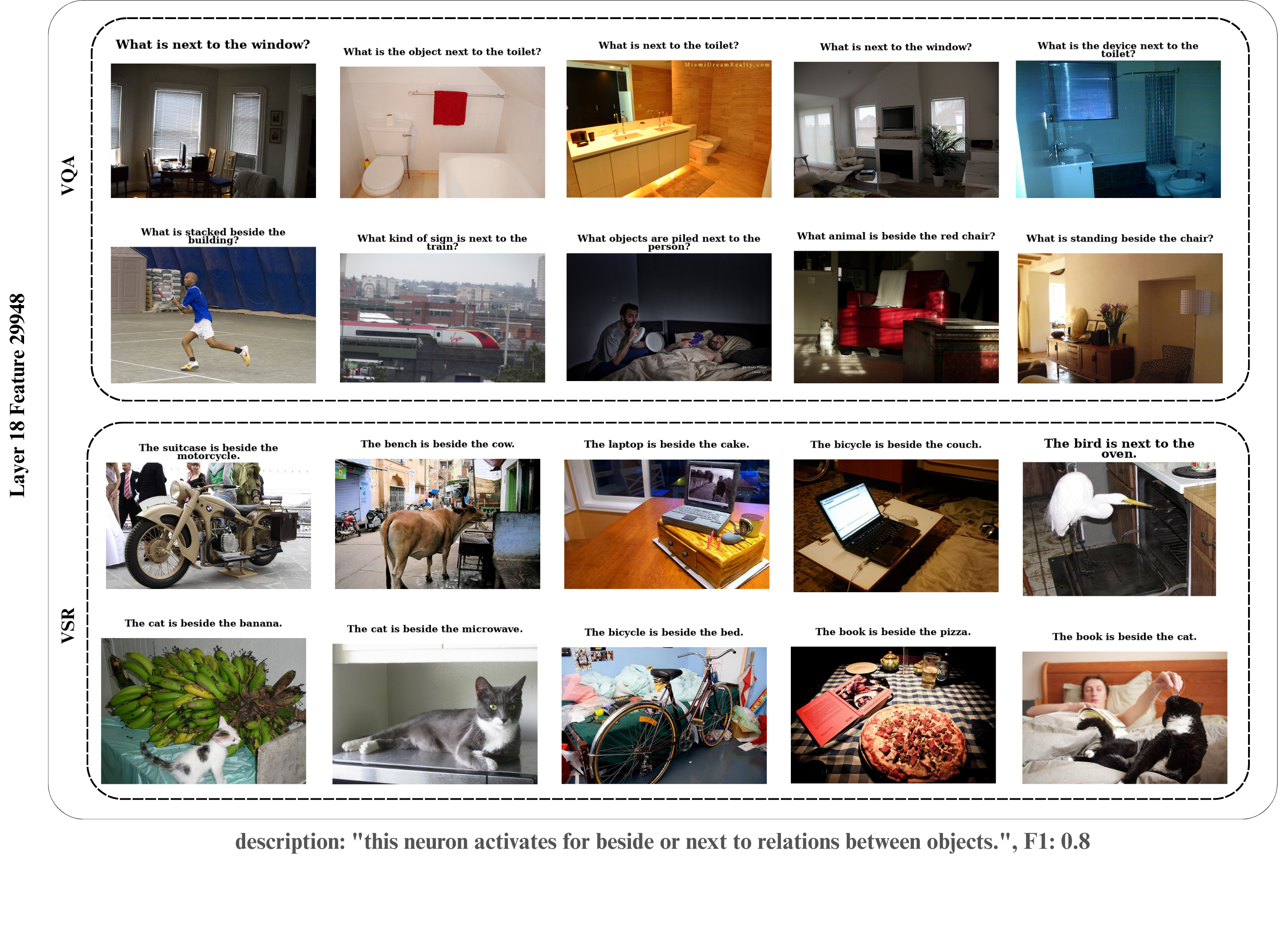}
    \label{fig:auto-interp-f29948}
  \end{subfigure}

  \vspace{-1.0em}
  \begin{subfigure}[t]{0.9\linewidth}
    \centering
    \includegraphics[width=\linewidth]{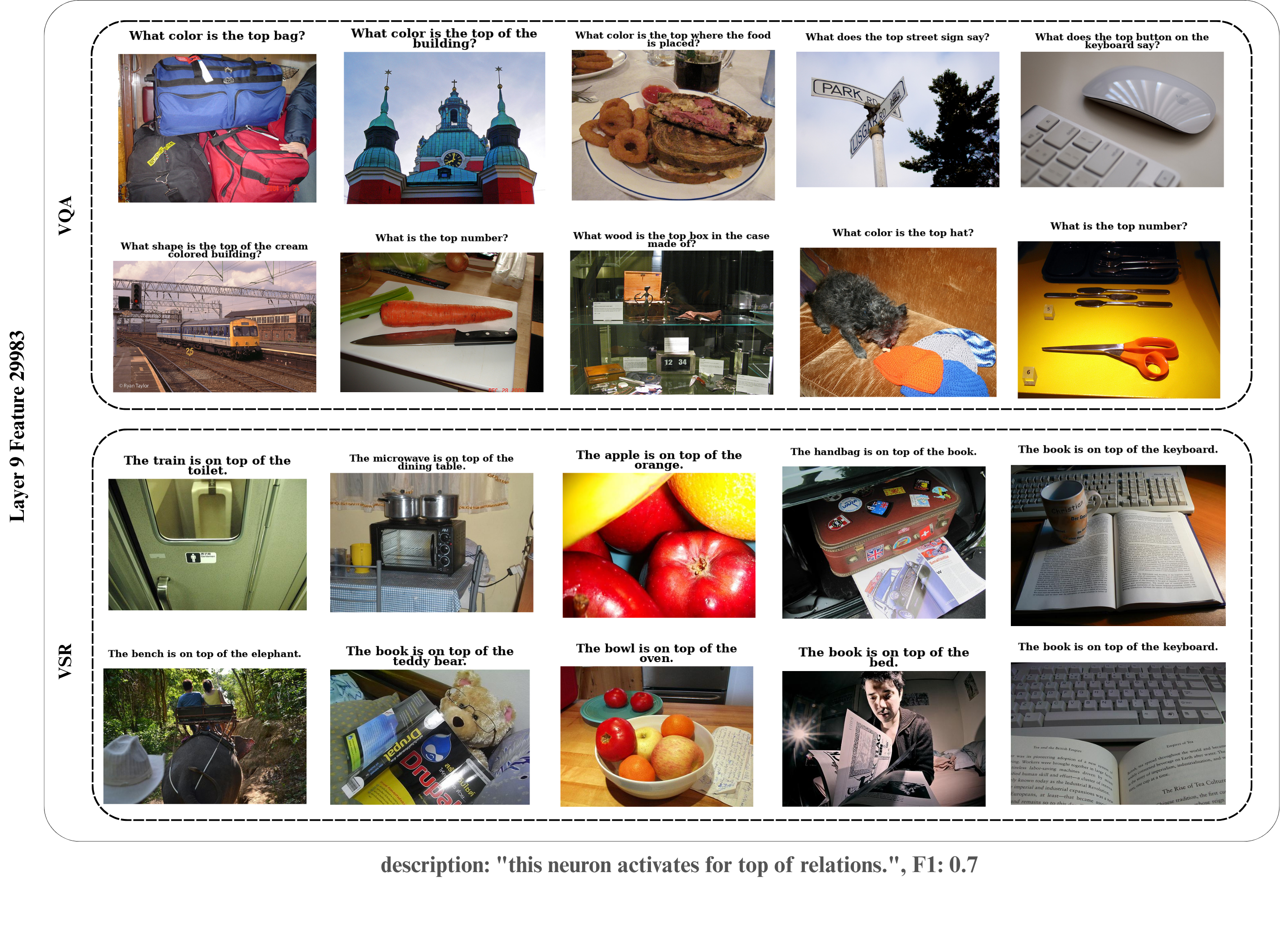}

    \label{fig:auto-interp-f29983}
  \end{subfigure}
  \vspace{-1em}
  \caption{\textbf{Additional Auto-Interp examples.} 
  Top-activating VQA and VSR samples for two adapted features, showing consistent spatial relations.}
  \label{fig:appendix-auto-interp}
\end{figure*}

\subsection{Auto-Interpretation and Scoring Pipeline}
\label{app:auto-interp}

We evaluate interpretability using an automated feature-description pipeline with two variants:
\emph{RAW} (image+text) and \emph{OVERLAY} (image+text+top-head heatmaps). For each feature $f$: 1. Select up to $k{=}5$ top-activating samples (deduped across VQA/VQA-spatial/VSR).
2. Call the API once to generate a single concise description.
3. Validate using held-out positive samples and random VQA negatives (two short rounds).
4. Compute F1 as a lightweight proxy for description confidence. Outputs are stored per feature as JSON (\texttt{description}, examples, classification results).
Adding overlays improves interpretability, with early results showing a typical gain of about $+0.2$ F1.

\begin{tcolorbox}[promptbox={PROMPT A: Description (RAW / OVERLAY)}]
\textbf{System:} You are analyzing individual neurons using their top-activating samples (image+text; OVERLAY also includes attention heatmaps).\\
\textbf{Task:} Produce \emph{one} short, lower-case sentence completing: ``this neuron activates for \ldots''.\\
\textbf{Guidelines:} Base it on consistent patterns supported by image (+ overlays) and text; be specific; no hedging.\\
\textbf{Return:} \{\,"description": "one concise sentence"\,\}.
\end{tcolorbox}

\vspace{-1em}
\begin{tcolorbox}[promptbox={PROMPT B: Validation (F1)}]
\textbf{System:} You are validating a neuron description against short examples (image+text; OVERLAY adds heatmaps).\\
\textbf{Task:} For each sample, output 1 if it reasonably matches the description; else 0.\\
\textbf{Return:} \{\,"classifications": [0/1, \dots]\,\}.
\end{tcolorbox}

\clearpage
% \newpage
\subsection{Attribution Patching Additional Experiments Results}

\subsection{Aggregated Attribution Results}
\label{app:attr-agg}

% \vspace{-1em}

\begin{figure*}[t]
    \centering
    \includegraphics[width=0.42\textwidth]{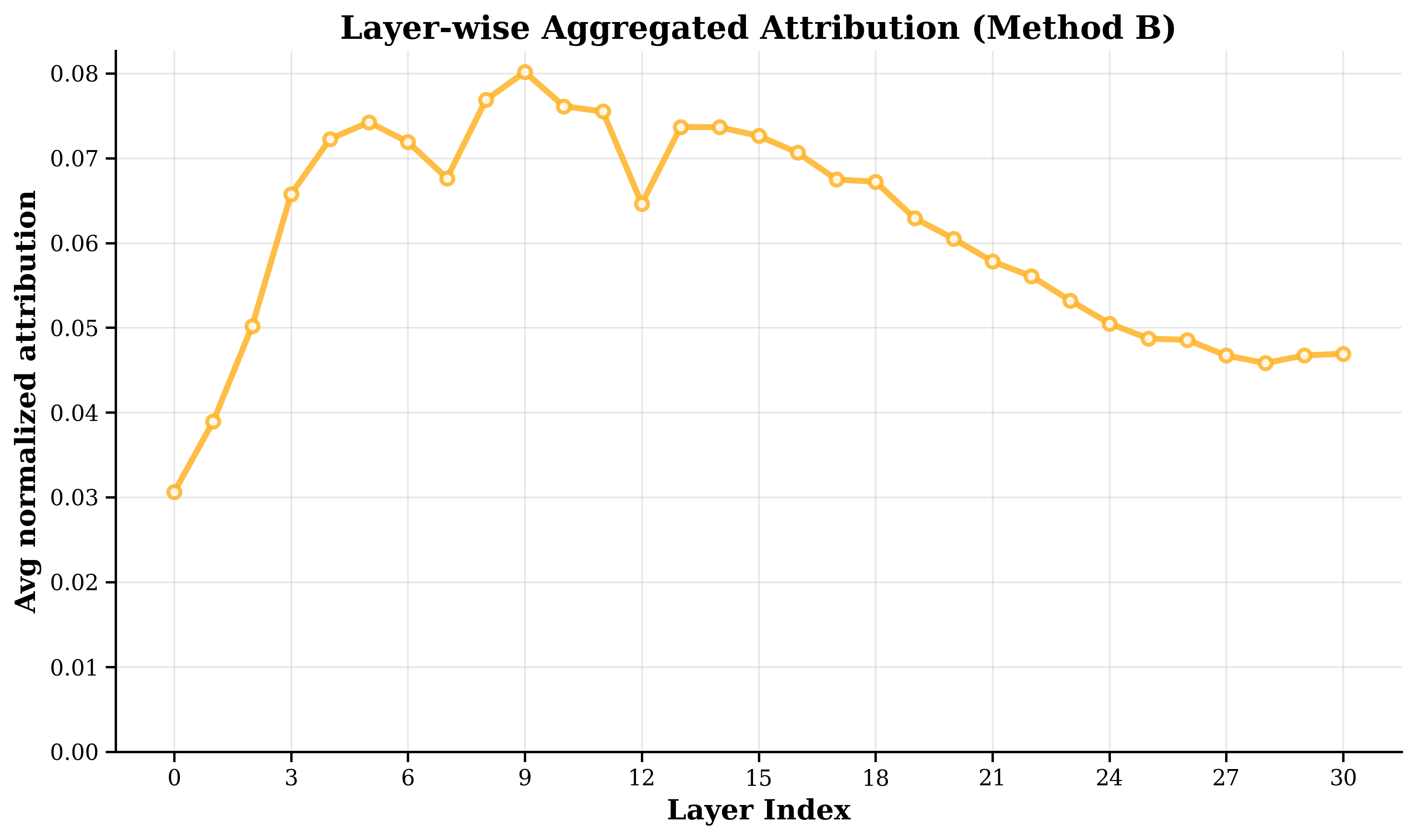}
    \hspace{1em}\includegraphics[width=0.42\textwidth]{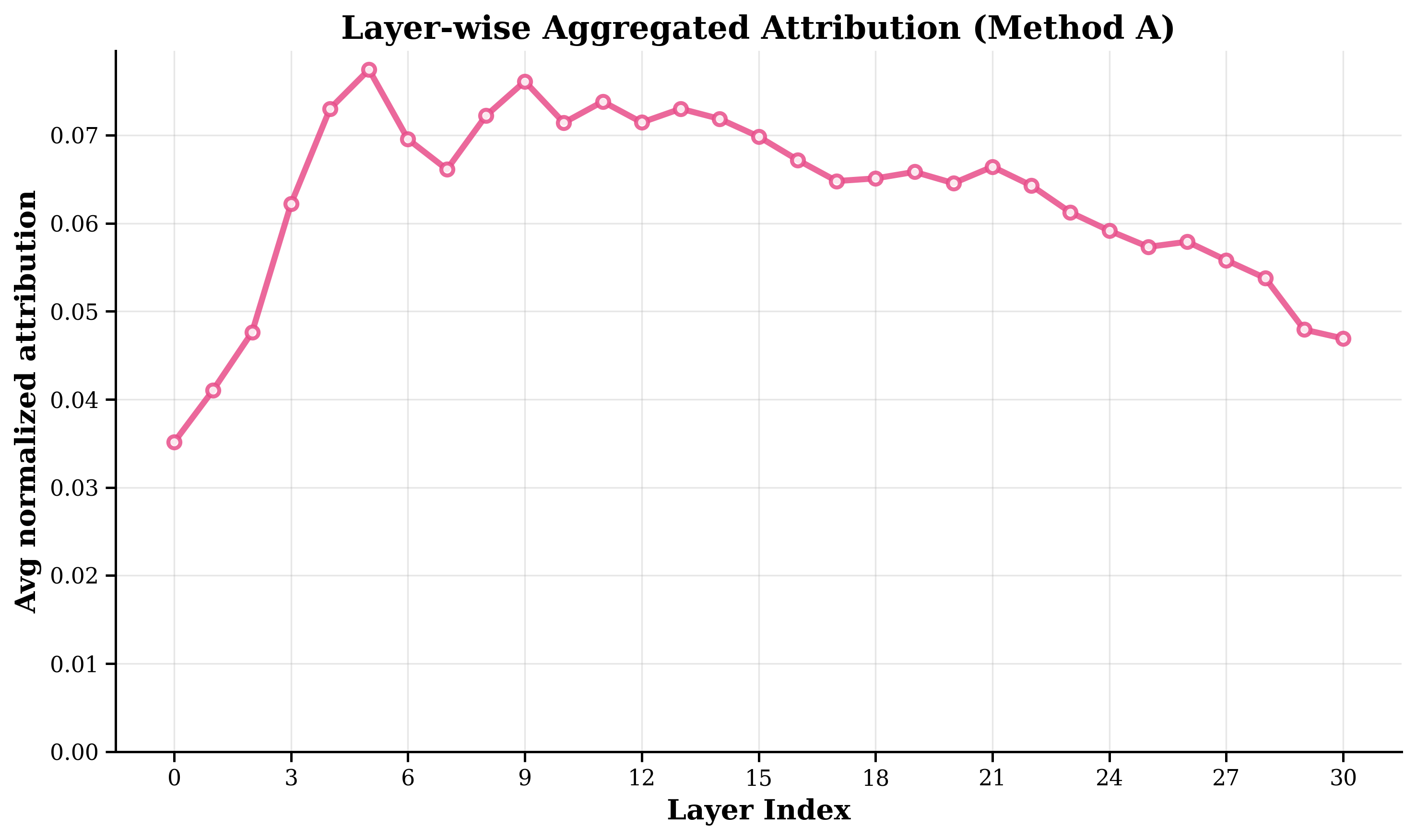}
    \caption{Layer-wise aggregated attribution curves for Method B (left) and Method A (right). Both peak in around middle layers, consistent with the emergence of spatial features.}
    \label{fig:layer-agg}
\end{figure*}
% \vspace{-2em}
\begin{figure*}[t]
    \centering
    \includegraphics[width=0.5\textwidth]{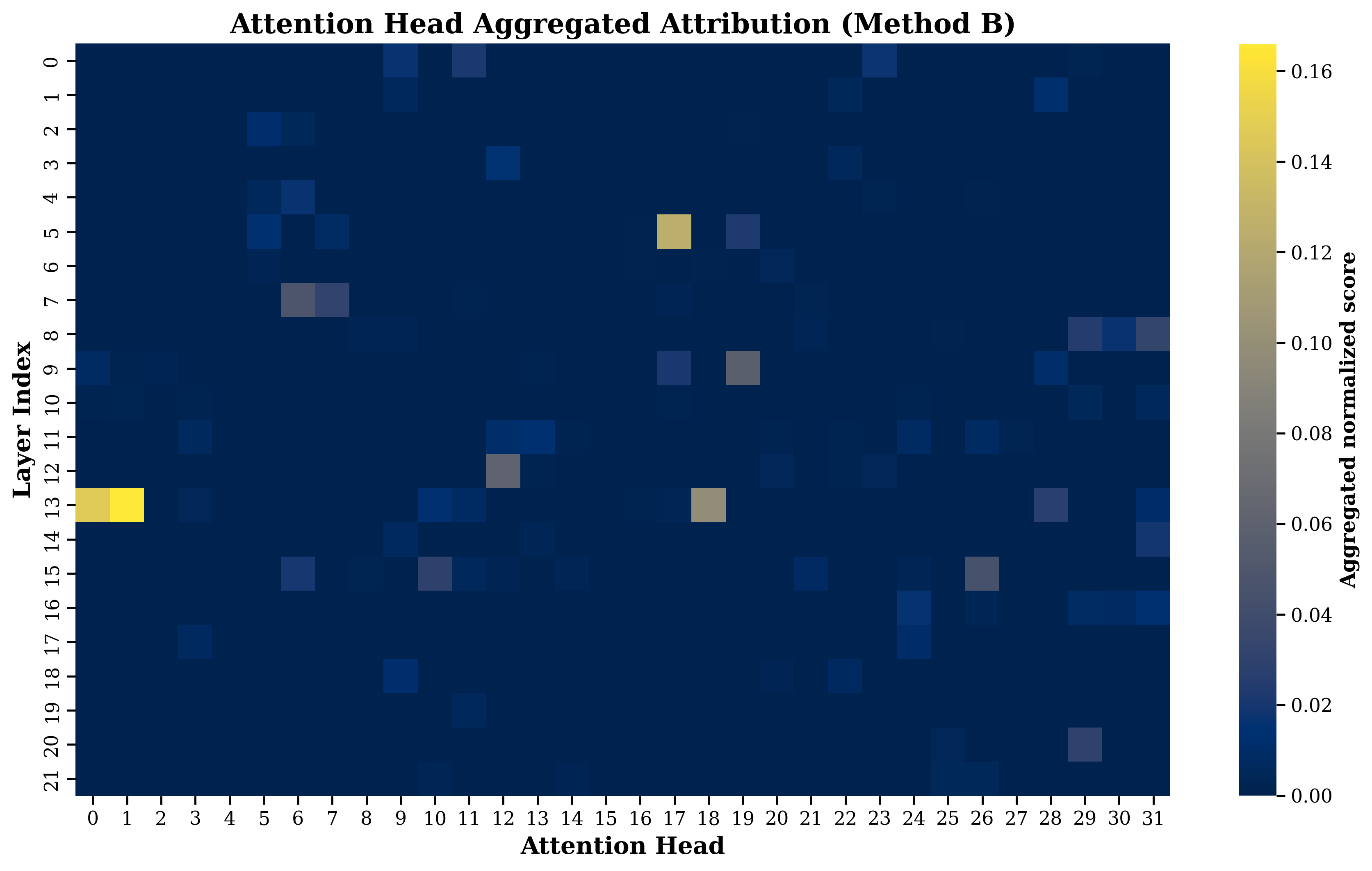}
    \includegraphics[width=0.45\textwidth]{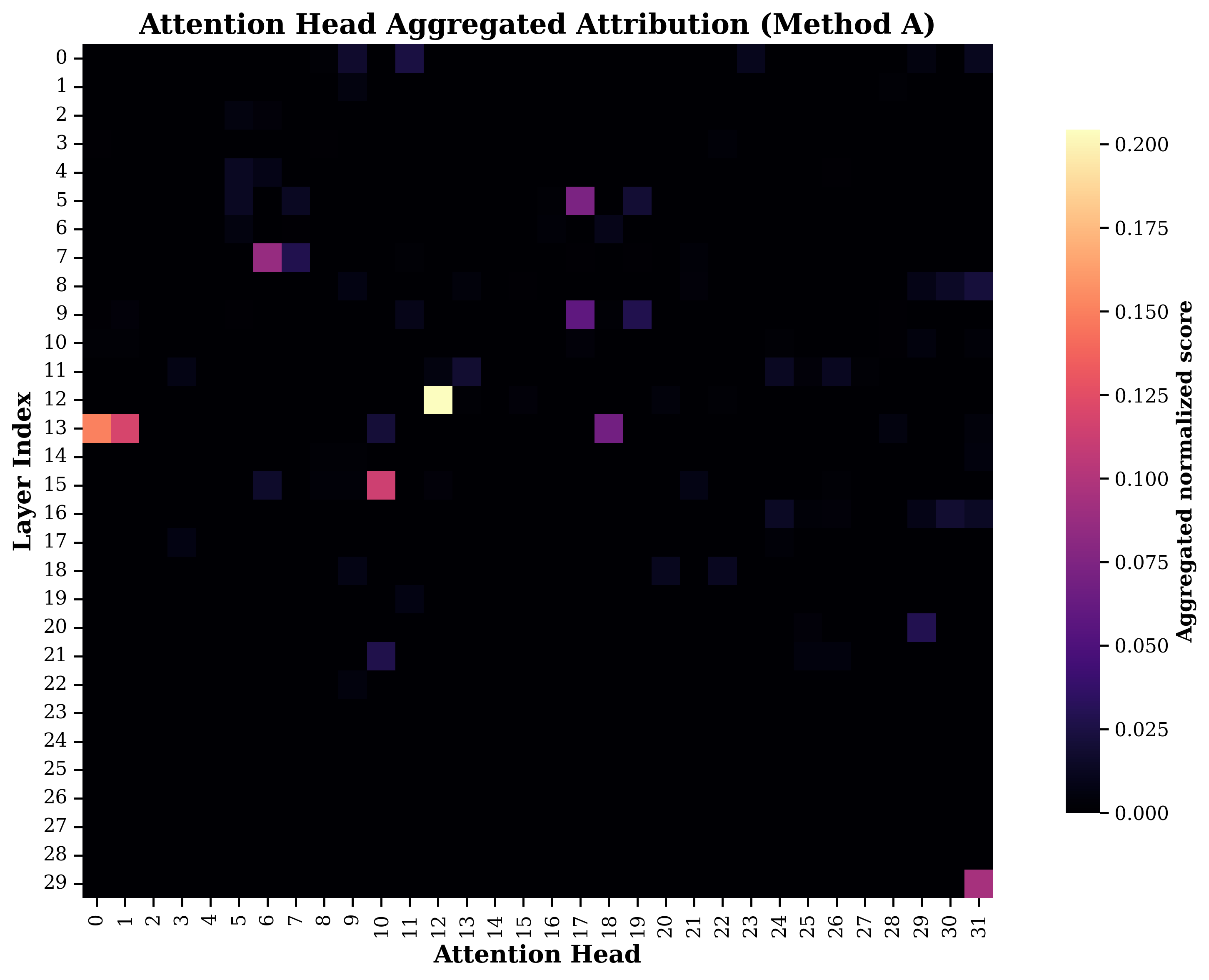}
    \caption{Attention head aggregated attribution maps for Method B (left) and Method A (right). Both highlight a similar set of specialized heads with high attribution scores.}
    \label{fig:head-agg}
\end{figure*}

\subsection{Per-Feature Panels with Top Heads}
\label{app:per-feature-panels}

For individual spatial features, we show (i) per-layer/head attribution maps (Methods A and B) and
(ii) attention overlays from the strongest heads on the feature’s top-activating samples across both
VSR and VQA datasets.

\begin{figure*}[t]
  \centering
  % ---- Feature panel: Layer 15, Feature 10748 ----
  \begin{subfigure}[t]{0.98\textwidth}
    \centering
    \includegraphics[width=\linewidth]{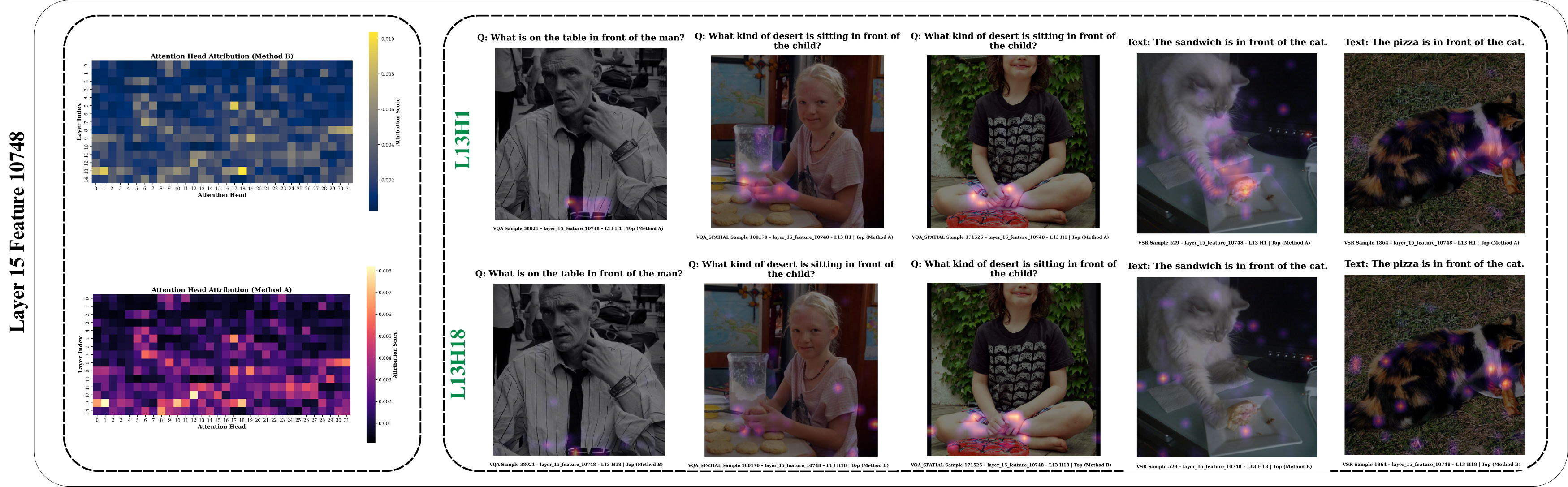}
    \caption{\textbf{Layer 15, Feature 10748.} VSR Relation: ``in front of.''\;
    Top heads (Method A): \texttt{L13H1}, \texttt{L12H12}, \texttt{L13H18}.\;
    Top heads (Method B): \texttt{L13H18}, \texttt{L5H17}, \texttt{L13H1}.\;
    \emph{Overlap}: \texttt{L13H1}, \texttt{L13H18}.
    Attention overlays are shown on the top-activating samples across VSR and VQA.}
    \label{fig:ap-single-l15f10748}
  \end{subfigure}

  % ---- Feature panel: Layer 20, Feature 22247 ----
  \begin{subfigure}[t]{0.98\textwidth}
    \centering
    \includegraphics[width=\linewidth]{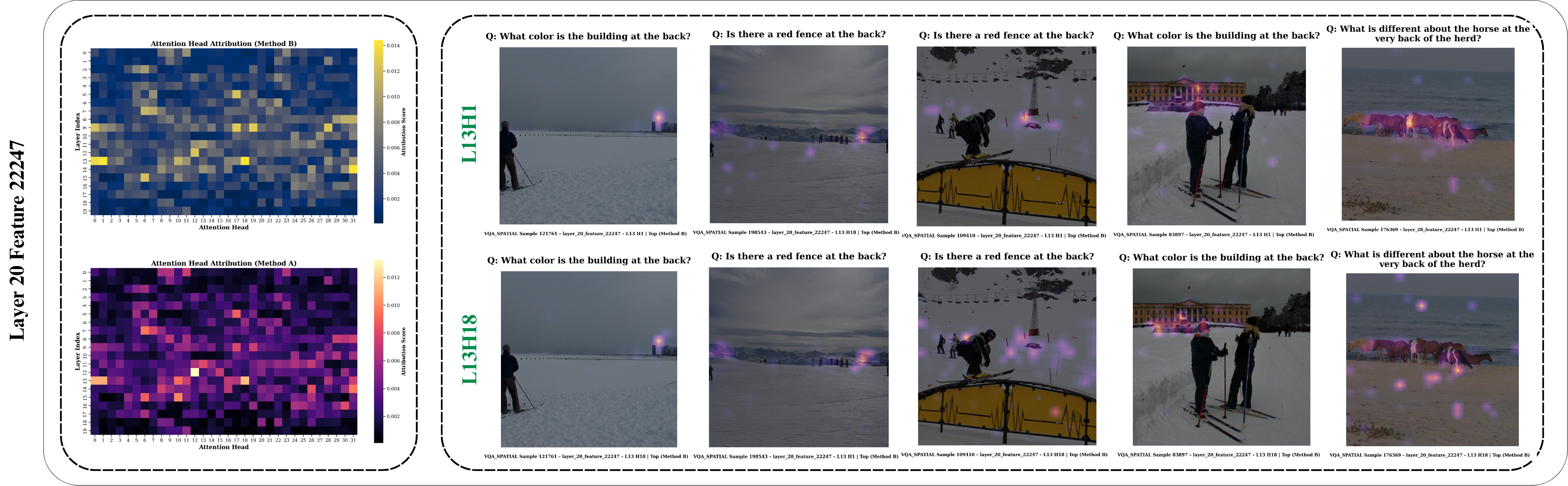}
    \caption{\textbf{Layer 20, Feature 22247.} VSR Relation: ``at the back of.''\;
    Top heads (Method A): \texttt{L12H12}, \texttt{L13H18}, \texttt{L13H1}.\;
    Top heads (Method B): \texttt{L13H1}, \texttt{L13H18}, \texttt{L14H31}.\;
    \emph{Overlap}: \texttt{L13H1}, \texttt{L13H18}.
    Attention overlays are shown on the top-activating samples across VSR and VQA.}
    \label{fig:ap-single-l20f22247}
  \end{subfigure}

  % ---- Feature panel: Layer 7, Feature 15870 ----
  \begin{subfigure}[t]{0.98\textwidth}
    \centering
    \includegraphics[width=\linewidth]{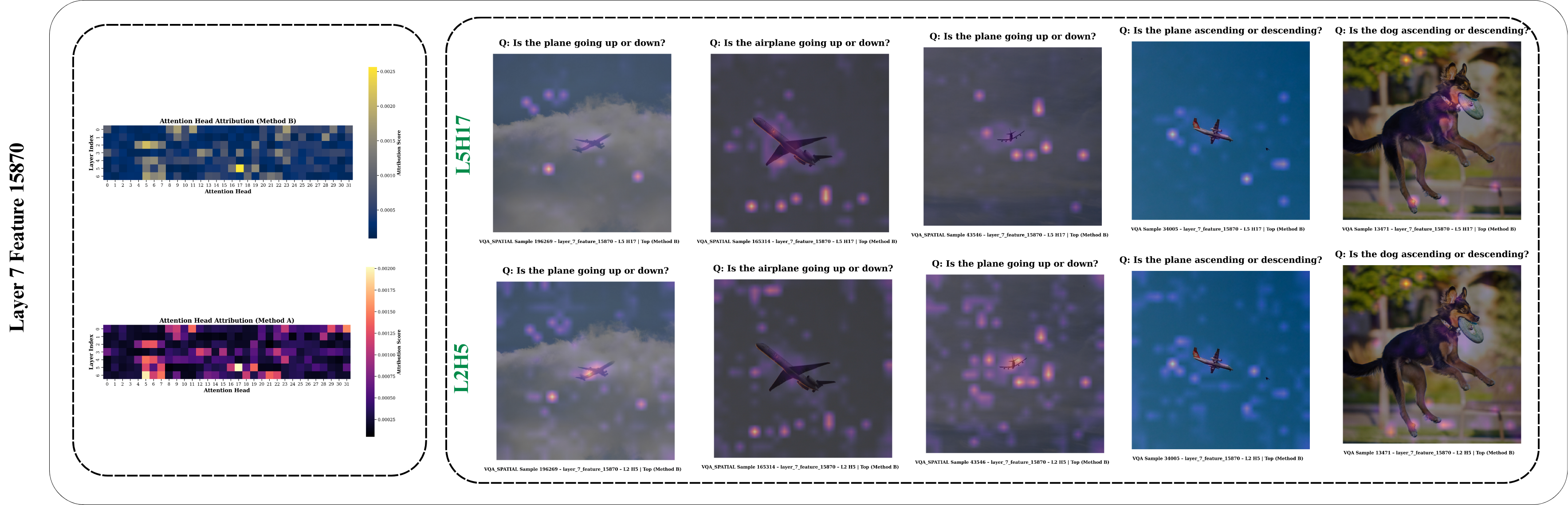}
    \caption{\textbf{Layer 7, Feature 15870.} VSR Relation: ``above.''\;
    Top heads (Method A): \texttt{L5H17}, \texttt{L6H5}, \texttt{L0H31}.\;
    Top heads (Method B): \texttt{L5H17}, \texttt{L2H5}, \texttt{L2H6}.\;
    \emph{Overlap}: \texttt{L5H17}.
    Attention overlays are shown on the top-activating samples across VSR and VQA.}
    \label{fig:ap-single-l7f15870}
  \end{subfigure}

  \caption{\textbf{Attribution patching on individual spatial features.}
  Each subfigure displays aggregated head/layer attribution maps (left) and attention overlays (right)
  using the strongest heads on the feature’s top-activating samples across both VSR and VQA.}
  \label{fig:ap-single-all}
\end{figure*}

Across these examples, the two attribution methods consistently surface overlapping heads, indicating that a small group concentrates much of the spatial signal. Method~B generally produces sharper rankings and cleaner overlays, suggesting it is more reliable for identifying the causal drivers of spatial features.

\subsection{Bottom-Ranked Heads as a Control}
\label{app:bottom-heads}

As a control, we visualize overlays from the \emph{bottom-ranked} heads (per method, per feature).  
Across VSR and VQA top-activating samples, these heads generally fail to localize semantically relevant regions.

\begin{figure*}[t]
  \centering
  % ---- Feature panel: Layer 15, Feature 10748 (bottom heads) ----
  \begin{subfigure}[t]{0.98\textwidth}
    \centering
    \includegraphics[width=\linewidth]{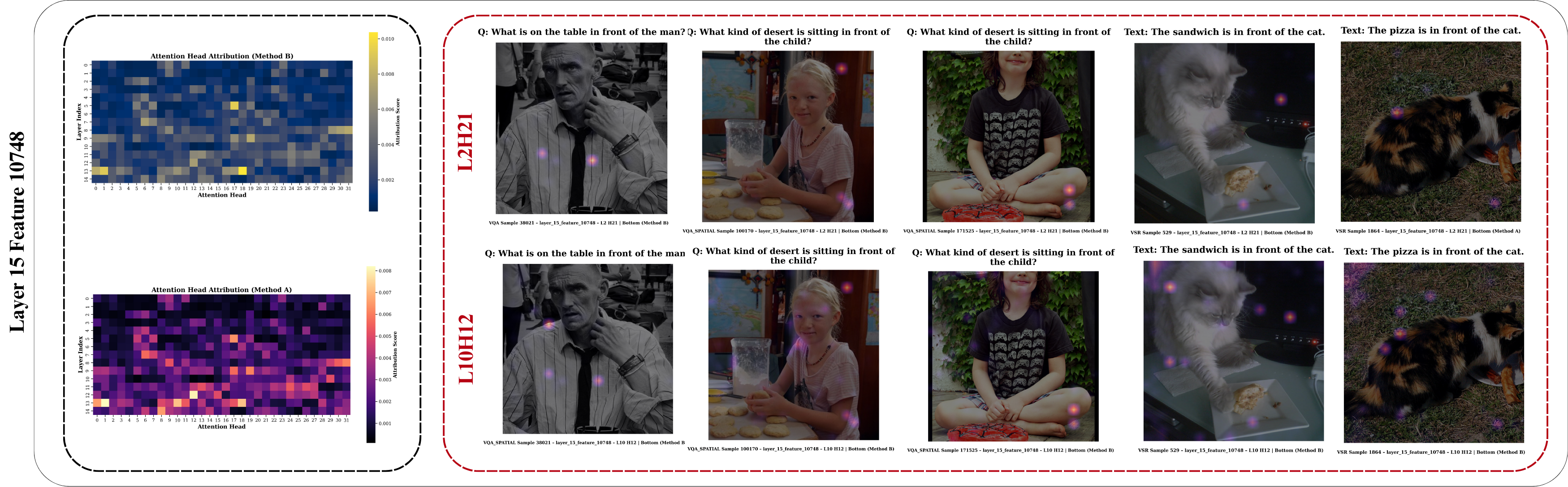}
    \caption{\textbf{Layer 15, Feature 10748.} VSR Relation: ``in front of.''}
    \label{fig:ap-neg-l15f10748}
  \end{subfigure}

  \vspace{1em}

  % ---- Feature panel: Layer 20, Feature 22247 (bottom heads) ----
  \begin{subfigure}[t]{0.98\textwidth}
    \centering
    \includegraphics[width=\linewidth]{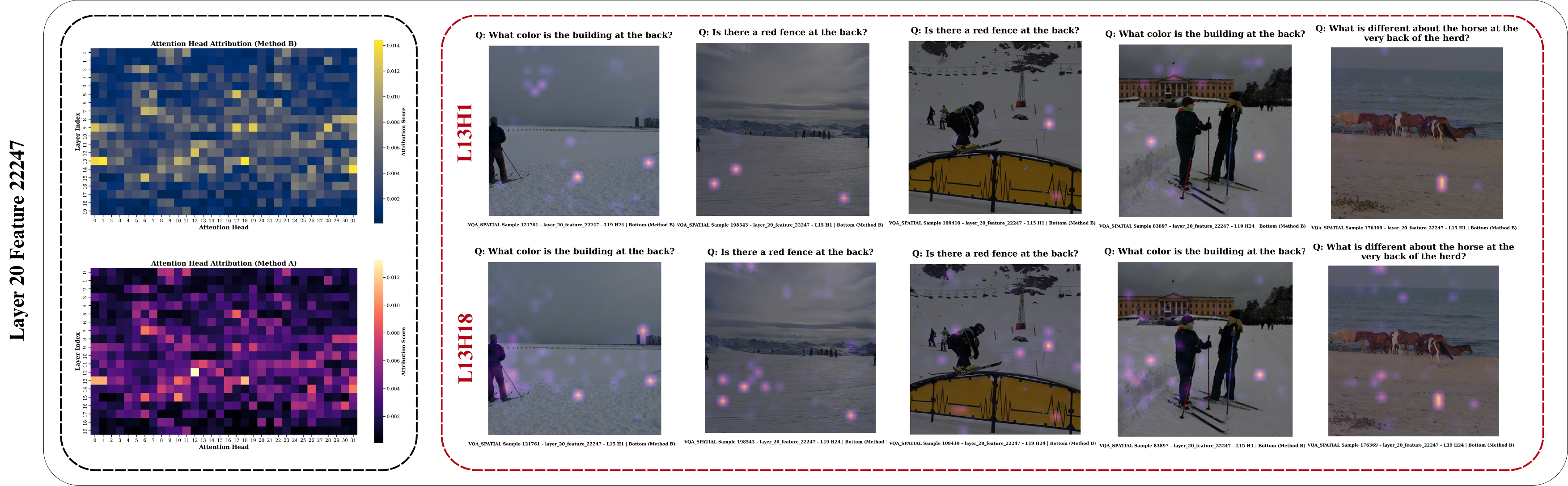}
    \caption{\textbf{Layer 20, Feature 22247.} VSR Relation: ``at the back of.''}
    \label{fig:ap-neg-l20f22247}
  \end{subfigure}

  \vspace{1em}

  % ---- Feature panel: Layer 7, Feature 15870 (bottom heads) ----
  \begin{subfigure}[t]{0.98\textwidth}
    \centering
    \includegraphics[width=\linewidth]{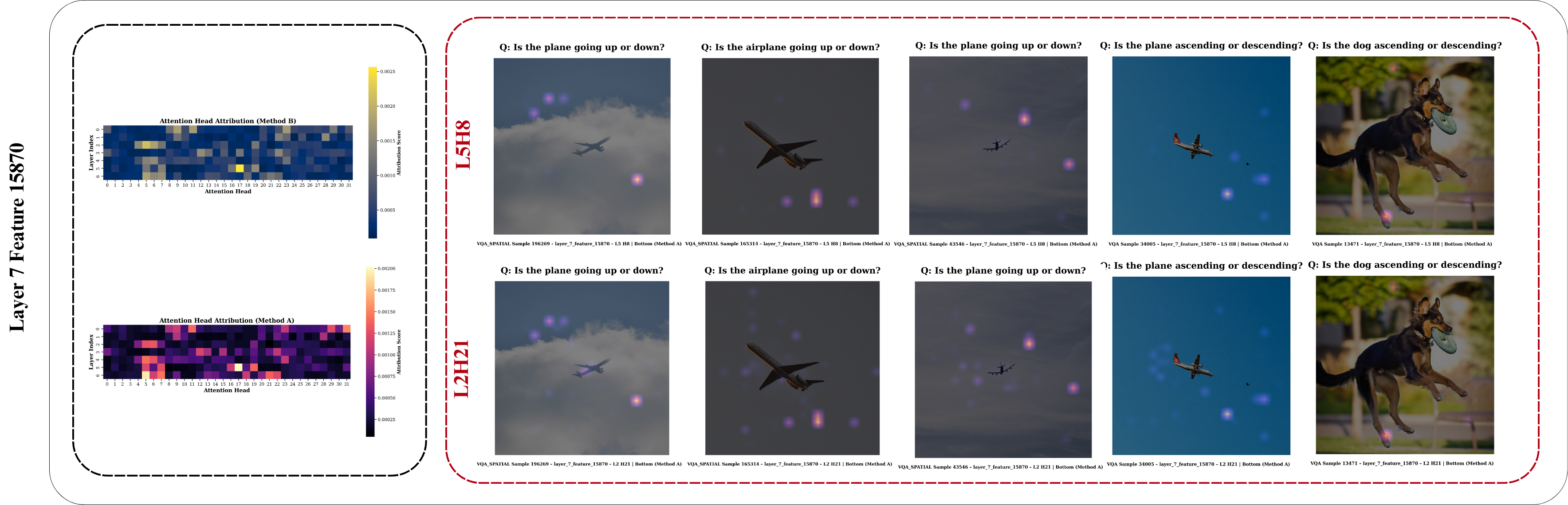}
    \caption{\textbf{Layer 7, Feature 15870.} VSR Relation: ``above.''}
    \label{fig:ap-neg-l7f15870}
  \end{subfigure}

  \caption{\textbf{Bottom-ranked heads yield weak localization.}
  For each feature, we show overlays from the lowest-scoring heads under Methods A and B on the feature’s
  top-activating samples across VSR and VQA. In contrast to Appx.~Fig.~\ref{fig:ap-single-all}, these
  heads produce diffuse or irrelevant attention.}
  \label{fig:ap-single-neg-all}
\end{figure*}

\subsection{Full Ablation Results}
\label{app:full-ablation}
Table~\ref{tab:full-ablation} reports a more detailed version of ablation results for the top SAE features. 
For each feature, we show average accuracy and probability drops on VSR across seeds, together with the number of evaluation samples. 
We also report accuracy drops on VQA, random-feature control drops ($\Delta$Ctrl), odds ratios (VSR OR), and relation-specific subsets of VSR derived from top-activating samples. 
Large negative $\Delta$VSR Acc with small $\Delta$VQA Acc indicates spatial specificity, near-zero $\Delta$Ctrl supports robustness, and high odds ratios reflect selective recruitment under spatial prompts. 
% \textit{Both VSR subsets and the VQAv2 Yes/No split are balanced binary classification; random guess is 50\%.}

\begin{table}[htbp]
\centering
\scriptsize
\setlength{\tabcolsep}{2.5pt}
\renewcommand{\arraystretch}{1.05}
\resizebox{\columnwidth}{!}{%
\begin{tabular}{
c c
S[table-format=-2.2]
S[table-format=-1.2]
S[table-format=-1.2]
S[table-format=-1.2]
S[table-format=1.2]
c
p{2.2cm}
}
\toprule
Layer & Feature & {$\Delta$VSR Acc} & {$\Delta$VSR Prob} & {$\Delta$VQA Acc} & {$\Delta$Ctrl} & {VSR OR} & {\#Samples} & VSR Relations \\
\midrule
11 & 27061 & -13.30 & -0.09 & -0.40 &  0.00 & 8.03 & 94  & across from \\
12 & 23874 & -10.24 & -0.10 & -0.40 & -0.95 & 9.10 & 210 & left of \\
18 & 29948 &  -7.98 & -0.09 & -0.30 &  0.00 & 8.36 & 188 & beside \\
23 & 4060  &  -5.85 & -0.00 & -0.70 & -1.06 & 7.47 & 94  & at the back of \\
14 & 17873 & -10.00 & -0.07 & -0.30 & -2.71 & 7.17 & 480 & at the right side of \\
9  & 15404 & -11.19 & -0.07 & -0.80 &  1.08 & 5.60 & 277 & below \\
7  & 6986  & -10.87 & -0.03 & -0.50 &  0.34 & 4.74 & 589 & under \\
7  & 15870 & -15.54 & -0.09 & -0.10 & -0.88 & 4.32 & 341 & above \\
10 & 5121  &  -7.92 & -0.06 & -0.10 &  0.12 & 4.22 & 846 & above, on top of \\
11 & 24089 &  -7.68 & -0.05 & -0.60 & -0.12 & 4.18 & 846 & above, on top of \\
12 & 13305 &  -6.38 & -0.05 & -0.70 &  0.24 & 4.17 & 846 & above, on top of \\
\bottomrule
\end{tabular}
}
\caption{Full ablation results for top SAE features, averaged over seeds.
The number of VSR samples evaluated is shown alongside accuracy/probability drops and odds ratios.}
\label{tab:full-ablation}
\end{table}

\begin{figure*}[h]
  \centering
    \centering
    % First grid
    \includegraphics[width=\linewidth]{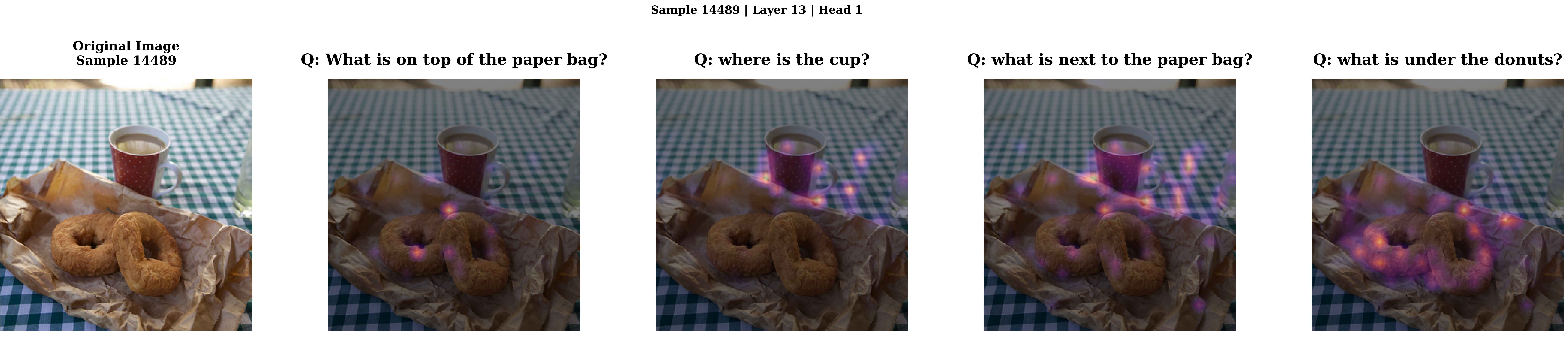}

    \vspace{0.5em} % small space between
    % Second grid
    \includegraphics[width=\linewidth]{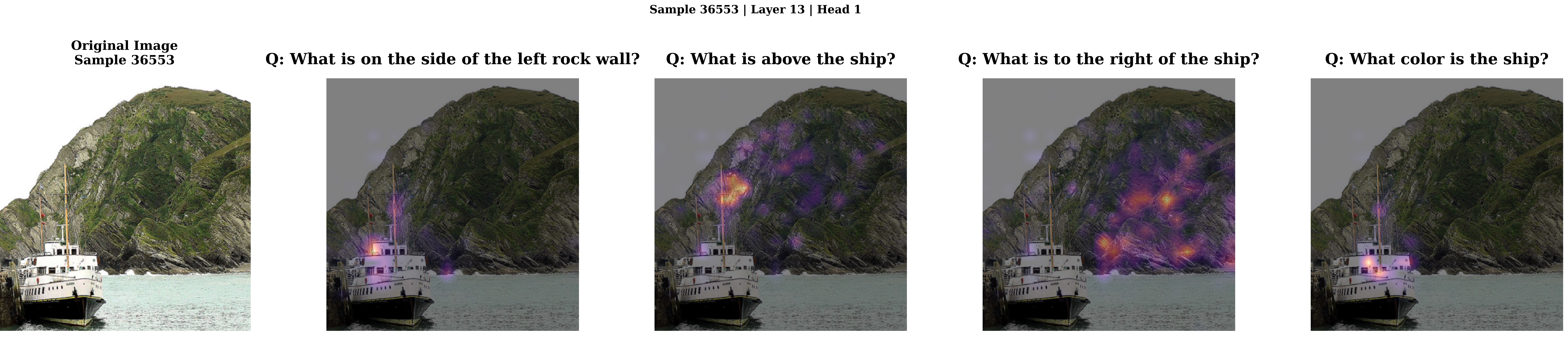}

    \vspace{0.5em}
    % Third grid
    \includegraphics[width=\linewidth]{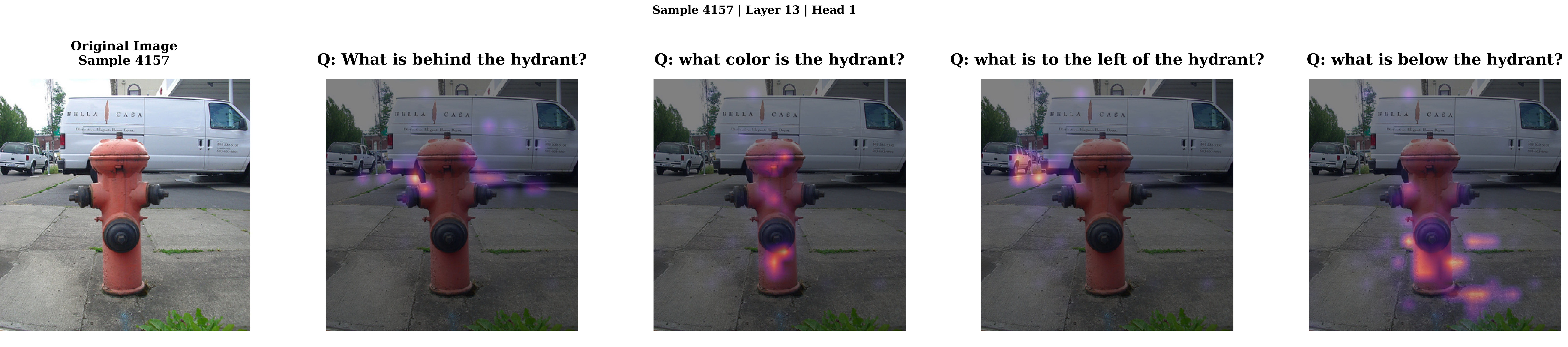}
  \caption{\textbf{Attention head visualizations across queries.} 
Each row shows one image with attention overlays from a single high-attribution head across multiple spatial and non-spatial custom queries. The same heads consistently focus on semantically relevant regions.}

  \label{fig:semantic-heads}
\end{figure*}

% \bibliographystyle{icml2026}
% \bibliography{references}

\end{document}